\documentclass[10pt,twocolumn,letterpaper]{article}

\usepackage{cvpr}
\usepackage[table,x11names,dvipsnames,table]{xcolor}
\usepackage{times}
\usepackage{epsfig}
\usepackage{graphicx}
\usepackage{amsmath}
\usepackage{amssymb}
\usepackage{pdfpages}
\usepackage{tabularx}

\usepackage{changes}

\usepackage{array}
\usepackage{booktabs}
\usepackage[pagebackref=true,breaklinks=true,letterpaper=true,colorlinks,bookmarks=false]{hyperref}

\definecolor{rowblue}{RGB}{220,230,240}
\definecolor{myorchid}{RGB}{150,10,30}
\definecolor{myblue}{RGB}{10,30,250}
\definecolor{mygreen}{RGB}{10,120,10}
\definecolor{myred}{RGB}{150,30,10}

\newcolumntype{R}[1]{>{\raggedleft\arraybackslash}p{#1}}
\newcolumntype{L}[1]{>{\raggedright\arraybackslash}p{#1}}
\newcolumntype{C}[1]{>{\centering\let\newline\\\arraybackslash\hspace{0pt}}m{#1}}

\cvprfinalcopy 

\def\cvprPaperID{5035} 

\ifcvprfinal\fi
\begin{document}

\title{Context-Aware Group Captioning via Self-Attention and Contrastive Features
}

\newcommand*{\affaddr}[1]{#1} 
\newcommand*{\affmark}[1][*]{\textsuperscript{#1}}
\newcommand*{\email}[1]{\texttt{#1}}

\author{%
Zhuowan Li\affmark[1]\thanks{\noindent This work has been done during the first author's internship at Adobe.}, Quan Tran\affmark[2], Long Mai\affmark[2], Zhe Lin\affmark[2], and Alan Yuille\affmark[1]\\
\affaddr{\affmark[1]Johns Hopkins University    }
\affaddr{\affmark[2]Adobe Research}\\
\email{\small \{zli110, alan.yuille\}@jhu.edu    }
\email{\small \{qtran, malong, zlin\}@adobe.com}
}

\maketitle

\begin{abstract}
   While image captioning has progressed rapidly, existing works focus mainly on describing single images. In this paper, we introduce a new task, context-aware group captioning, which aims to describe a group of target images in the context of another group of related reference images. Context-aware group captioning requires not only summarizing information from both the target and reference image group but also contrasting between them. To solve this problem, we propose a framework combining self-attention mechanism with contrastive feature construction to effectively summarize common information from each image group while capturing discriminative information between them. To build the dataset for this task, we propose to group the images and generate the group captions based on single image captions using scene graphs matching. Our datasets are constructed on top of the public Conceptual Captions dataset and our new Stock Captions dataset. Experiments on the two datasets show the effectiveness of our method on this new task. \footnote{Related Datasets and code are released at \url{https://lizw14.github.io/project/groupcap}.}
   
\end{abstract}

\section{Introduction}

\begin{figure}[h!]
\begin{center}
\includegraphics[width=0.45\textwidth]{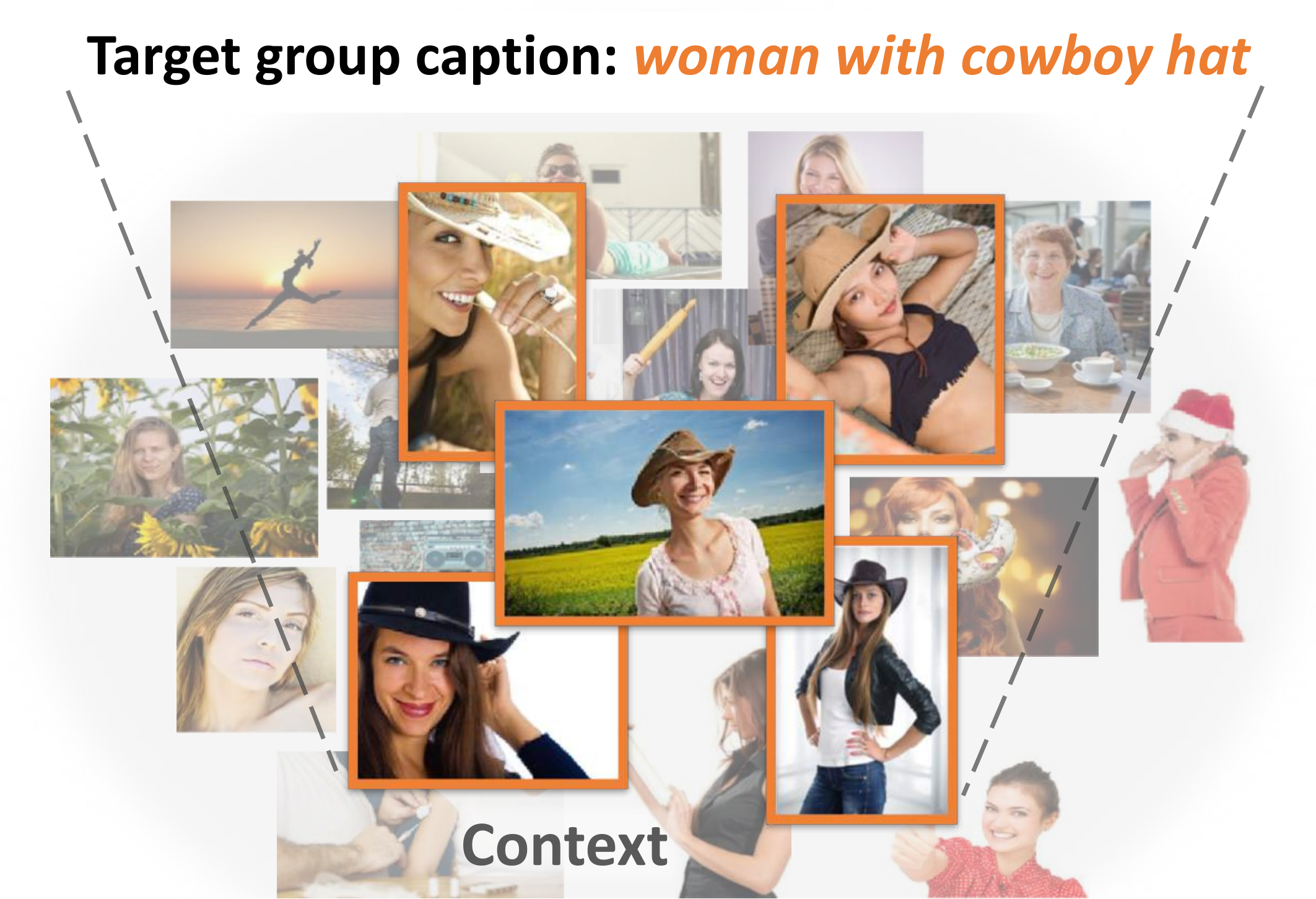}
\end{center}
   \caption{Context-ware group captioning. Given a group of target images (shown in orange boxes) and a group of reference images which provide the context (\texttt{woman}), the goal is to generate a language description (\texttt{woman with cowboy hat}) that best describes the target group while taking into account the context depicted by the reference group.} \vspace{-0.1in}
\label{fig:introduction}
\end{figure}



Generating natural language descriptions from images, the task commonly known as image captioning, has long been an important problem in computer vision research~\cite{bai2018survey, hossain2019comprehensive, liu2019survey}.
It requires a high level of understanding from both language and vision. Image captioning has attracted a lot of research attention in recent years thanks to the advances in joint language-vision understanding models~\cite{anderson2018bottom, karpathy2015deep, rennie2017self, xu2015show}. While image captioning has progressed rapidly, existing works mostly focus on describing individual images. There are practical scenarios in which captioning images in group is desirable. Examples include summarizing personal photo albums for social sharing or understanding web user intention from their viewed or clicked images. Moreover, it is often the case that the target image group to be captioned naturally belongs to a larger set that provides the context. For instance, in text-based image retrieval applications, given a group of user-interested images and other images returned by the search engine, we could predict the user hidden preferences by contrasting the two groups and suggest a new search query accordingly. Figure~\ref{fig:introduction} shows an example of such scenario. Among all the images returned by search query \texttt{woman}, the user can indicate his/her interest in some of the images (in orange boxes). The objective is to recognize that the user wants \texttt{woman with cowboy hat} and suggest the query accordingly.





Inspired by these real-world applications, we propose the novel problem of \textit{context-aware group captioning}: given a group of target images and a group of reference images, our goal is to generate a language description that best describes the target group in the context of the reference group. Compared to the conventional setting of single-image based captioning, our new problem poses two fundamental requirements. First, the captioning model needs to effectively summarize the common properties of the image groups. Second, the model needs to accurately describe the distinguishing content in the target images compared to the reference images.

To address those requirements, we develop a learning-based framework for context-aware image group captioning based on self-attention and contrastive feature construction. To obtain the feature that effectively summarizes the visual information from the image group, we develop a group-wise feature aggregation module based on self-attention. To effectively leverage the contrastive information between the target image group and the reference images, we model the context information as the aggregated feature from the whole set and subtract it from each image group feature to explicitly encourage the resulting feature to capture the differentiating properties between the target image group and the reference image group.



Training our models requires a large number of image groups with text descriptions and associated reference image sets. In this paper, we leverage large-scale image caption datasets to construct the training data. In particular, we build our annotations on top of the Conceptual Captions~\cite{sharma2018conceptual}, a recently introduced large-scale image captioning dataset. We parse the single-image caption into scene graphs and use the shared scene graphs of image groups to generate the groups' ground-truth captions. In addition, we apply the same procedure on a large-scale image set collected from a photograph collection. This dataset contains a large number of images with compact and precise human-generated per-image descriptions. That results in our second dataset, \emph{Stock Captions}, which we plan to contribute to the research community to encourage future research in this new problem.  


Our main contributions in this paper are three-fold. First, we introduce the problem of context-aware group captioning. This novel image captioning setting can potentially be important for many real-world applications such as automatic query suggestion in image retrieval. Second, we present a learning-based approach which learns to aggregate image group visual feature for caption generation. Our framework combines the self-attention mechanism with contrastive feature construction to effectively encode the image group into a  context-aware  feature  representation, which effectively summarizes relevant common information in the groups while capturing discriminative information between the target and context group. Third, we introduce two large-scale datasets specifically for the context-aware group captioning problem. Experiments on the two datasets demonstrate that our model consistently outperforms various baselines on the context-based image group captioning task.

\section{Related Work}

{\bf Image captioning} has emerged as an important research topic with a rich literature in computer vision~\cite{bai2018survey, hossain2019comprehensive, liu2019survey}. With the advances in deep neural networks, state-of-the-art image captioning approaches \cite{anderson2018bottom, donahue2015long, jiang2018recurrent, karpathy2015deep, ordonez2011im2text, rennie2017self, vinyals2015show, yao2018exploring} are based on the combination of convolutional neural networks~\cite{krizhevsky2012imagenet} and recurrent neural networks~\cite{hochreiter1997long} (CNN-RNN) architecture, where the visual features are extracted from the input image using CNNs which is then decoded by RNNs to generate the language caption describing the given image. Research in image captioning has progressed rapidly in recent years. Novel network architectures~\cite{anderson2018bottom, chen2017sca, lu2017knowing,wang2016image}, loss functions~\cite{dai2017towards, liu2017improved, liu2018show, luo2018discriminability, rennie2017self, shetty2017speaking}, and advanced joint language-vision modeling techniques~\cite{johnson2016densecap, kim2019dense, lu2017knowing, xu2015show, yang2017dense, you2016image} have been developed to enable more diverse and discriminative captioning results. Recent works have also proposed to leverage the contextual and contrastive information from additional images to help generating more distinctive caption for the target image~\cite{andreas2016reasoning, chen2018groupcap,dai2017contrastive,vedantam2017context,huang2016visual} or comparative descriptions between image pairs~\cite{park2019robust,suhr2017corpus,suhr2018corpus,tan2019expressing}. Existing works, however, mostly focus on generating captions for a single image. Our work, on the other hand, focuses on the novel setting of context-based image group captioning which aims to describe a target image group while leveraging the context of a larger pool of reference images.


{\bf Referring expression generation}~\cite{kazemzadeh2014referitgame,luo2017comprehension,yu2017joint} is a related problem to image captioning, which aims to generate natural language descriptions for a target object in an image. Contrastive modeling has been successfully applied in state-of-the-art referring expression generation methods to describe the target image region in contrast with other image regions. Yu \etal~\cite{yu2016modeling} use relative location and feature difference to discriminate the target object. Mao \etal~\cite{mao2016generation} maximize the probability of generated expression describing a specific region over other regions by Maximum Mutual Information training. While referring expression generation considers the target region in contrast with each negative region respectively, our problem requires contrastive context modeling among and between image groups.

{\bf Attention mechanism} has been successful in image captioning \cite{chen2017sca,liu2017attention, lu2017knowing, xu2015show, you2016image}. 
These works focused on applying visual attention to different spatial regions at each text generation time step.
More recently, attention in transformer\cite{vaswani2017attention} and pretrained BERT\cite{devlin2018bert} has been very successful in natural language processing tasks. 
\cite{li2019visualbert, lu2019vilbert, sun2019videobert} adapts the idea of BERT to vision and language tasks and showed improved performance on multiple sub-tasks. \cite{wang2018non} bridges attention and non-local operator to capture long-range dependency, which has been used in many computer vision tasks \cite{zhu2019asymmetric, li2020neural, cao2019gcnet, yue2018compact}. 
In our work, we apply attention over a group of images and show its effectiveness for summarizing information in an image group.

Our setting is inspired by {\bf query suggestion}~\cite{dehghani2017learning, jiang2018rin, sordoni2015hierarchical, wu2018query} in the context of document retrieval systems. Query suggestion aims to predict the expanded query given the previous query used by the users while taking into account additional context such as search history~\cite{dehghani2017learning, jiang2018rin, sordoni2015hierarchical} or user interaction (e.g. clicked and skipped documents)~\cite{wu2018query}. We are inspired by this task formulation and extend it to vision domain. Earlier works on query suggestion in image search focus on forming visual descriptors to help obtain better search results~\cite{zha2009visual,zha2010_TOMM} while the suggested text query is obtained solely from the current user query without taking visual content understanding into account. Our work, on the other hand, can potentially be applied to enable query suggestion from images. In this work, we focus on the image captioning aspect without relying on modeling user information and behavior as in existing query suggestion works, thus making it applicable beyond retrieval tasks.



\section{Dataset} \label{sec:dataset}


To train our models, we need a large-scale dataset where each data sample contains a group of target images with an associated ground-truth description and a larger group of reference images. The reference images need to be relevant to target images while containing a larger variety of visual contents and thus provides context for describing target images. The description should be both specific to the target group and conditioned on the reference group.

In this section, we first describe the intuition and method for dataset creation, then provide details of our proposed datasets built on the Conceptual Captions dataset and our proposed Stock Captions dataset.

\subsection{Data Construction Method} \label{sec:data method}

\begin{figure}[t]
\begin{center}
\includegraphics[width=0.48\textwidth]{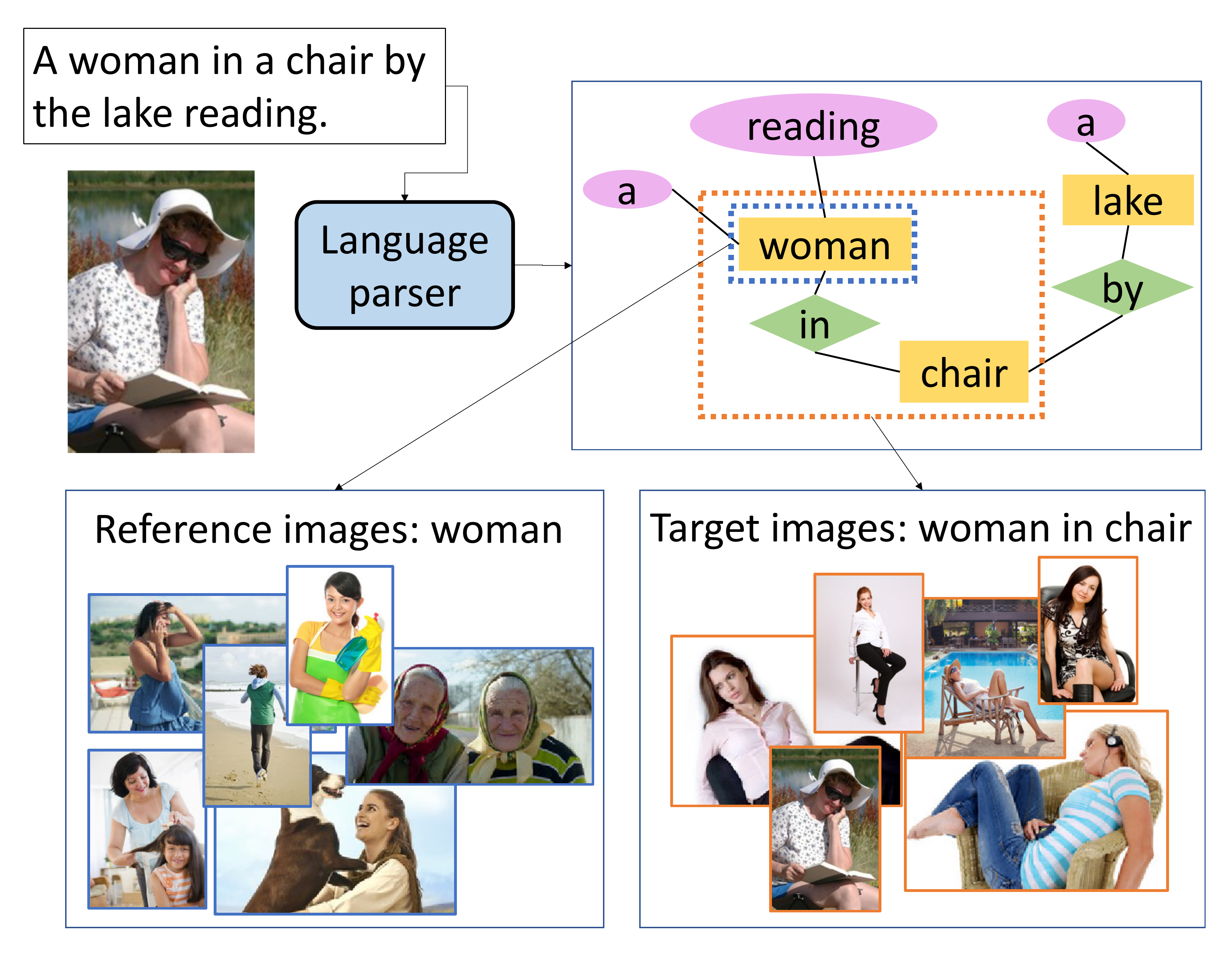}
\end{center}
   \caption{Dataset construction method. Our datasets are constructed from image collections with per-image descriptions. A pretrained language parser is used to parse each image caption into a scene graph. Then the images with shared scene graph are grouped to form the target group. Images with scene graphs that partially match the targets' form the reference group.}
\label{fig:data generation}
\end{figure}

We build our dataset on top of large-scale per-image captioning datasets by leveraging the shared scene graphs among images, motivated by \cite{chen2018groupcap}. The overall data generation process is shown in Figure \ref{fig:data generation}.

Images with shared scene graphs compose an image group.
More specifically, images with the same \emph{(attribute)-object-relationship-(attribute)-object} are chosen to compose the target image group, while images with partially overlapping scene graphs with the target group are chosen as the reference image group. 
For example, as in Figure \ref{fig:data generation}, images with the scene graph \emph{woman in chair} are selected to form the target group, while images containing \emph{woman} are selected to form the reference group paired with the target group. 
In this way, the reference group contains a larger variety of contents (woman in any places or poses) while the target group is more specific in terms of certain attributes (in chair).


In order to get the scene graphs for each image to support our grouping process, we use a pretrained language parser (improved upon \cite{wang2018scene}) to parse each ground-truth per-image caption into a scene graph. 
We choose to parse the scene graph from image captions instead of using the annotated scene graph in Visual Genome dataset~\cite{krishna2017visual}  because our scene graph needs to focus on the most "salient" content in the image.
Since Visual Genome is densely annotated without the information of which object is the main content of the image, scene graphs of small trivial objects may dominate the grouping process while the main content is ignored. This will produce very noisy data, potentially unsuitable for training our models. 
On the other hand, while parsing errors may introduce noise, scene graphs parsed out of image captions focus on the main objects because the caption usually describes the most important contents in an image.

After getting the target and reference groups using scene graph matching, the shared scene graph among target images is flattened into text to serve as the ground truth group description. For example, in Figure \ref{fig:data generation}, the ground-truth group caption is \emph{woman in chair}. 
Other examples of ground-truth group captions include:  \textit{colorful bag on white background, girl in red, business team holding terrestrial globe, woman with cowboy hat, etc. }

To construct our datasets for group captioning, the per-image captioning datasets need to be large-scale to provide enough image groups. We build our group captioning datasets on top of two datasets: Conceptual Captions dataset~\cite{sharma2018conceptual}, which is the largest existing public image captioning dataset, and Stock Captions dataset, which is our own large-scale per-image captioning dataset characterized by precise and compact descriptions. Details about construction on the two datasets are provided as follows.\footnote{For simplicity, in this paper, we call our newly constructed group captioning datasets by the same name as their parent datasets: Conceptual Captions, and Stock Captions.
}

\subsection{Conceptual Captions}
Conceptual Captions is a large-scale image captioning dataset containing 3.3 million image-caption pairs.
(By the time we download the images through the urls provided, only 2.8 million are valid.)
Because the captions are automatically collected from alt-text enabled images on the web, some of the captions are noisy and not natural.
However, the high diversity of image contents and large number of images makes Conceptual a suitable choice for data generation using our method.

After sampling from 2.7 million images from Conceptual Captions, we obtain around 200k samples with 1.6 million images included.
Each sample contains 5 target images and 15 reference images.
The images with rare scene graphs that cannot be made into groups are not used.
We manually cleaned the sampled data to remove samples that are not meaningful. For example, target group of \emph{portrait or woman} and reference group of \emph{woman} are not semantically different so they are removed. 
We also cleaned the vocabulary to remove rare words. 

The 200k samples are split into test, validation and train splits, where these three splits share the same image pool.
While the validation and train splits may contain samples of same group captions (because group captions are usually short), we make sure that captions in test split do not overlap with train split.
More detailed statistics are provided in Table \ref{tab:dataset}.


\begin{table}[b]
\small{
  \begin{center}
    \begin{tabular}{L{2cm} L{2cm} L{2cm}}
        
        \toprule
        \multicolumn{3}{c}{\textbf{Original Per-Image Captioning}}\\
         & {Conceptual} & {Stock}\\
        \specialrule{0.3pt}{1pt}{1pt}
        Size & 2766614 & 5785034 \\
        Avg Length & 9.43 & 4.12 \\
        \specialrule{1pt}{1pt}{1pt}
        \multicolumn{3}{c}{\textbf{Context-aware Group Captioning}} \\
         & {Conceptual} & {Stock}\\
        \specialrule{0.3pt}{1pt}{1pt}
        Size & 199442 & 146339 \\
        Train Split & 175896 & 117829 \\
        Val Split & 10000 & 10000 \\
        Test Split & 13546 & 18510 \\
        \# of images & 1634523 & 1941370 \\
        Vocab Size & 5811 & 2437 \\
        Avg Length & 3.74 & 2.96 \\
        \bottomrule
    \end{tabular}
  \end{center}
  \caption{Statistics of Conceptual Captions and Stock Captions, in terms of original per-image captioning dataset and our group captioning dataset constructed on top of per-image captions.}
  \label{tab:dataset}
 }
\end{table}

\subsection{Stock Captions}
While the Conceptual dataset excels in image diversity, we found that its captions are often long and sometime noisy.
Motivated by the query suggestion application where the suggested search queries are usually short and compact, we propose to construct the dataset on a new image captioning dataset named Stock Captions.
Stock Captions is a large-scale image captioning dataset collected in text-to-image retrieval setting. Stock Captions dataset is characterized by very precise, short and compact phrases. 
Many of the captions in this dataset are more attribute-like short image titles, e.g. "colorful bags", "happy couple on a beach", "Spaghetti with dried chilli and bacon", etc. 

After grouping and filtering the 5.8 million raw images, we get 1.9 million images, grouped into 1.5 million data samples for the Stock Captions dataset. The dataset sampling and split details are similar to Conceptual.(Table \ref{tab:dataset}).

\subsection{User Study for Dataset Comparisons}

To test the quality of our data and compare our two datasets,
we conduct a user study by randomly selecting 500 data samples (250
from each dataset) and ask 25 users to 
give a 0-5 score for each sample. 

\begin{figure}[t]
\begin{center}
\includegraphics[width=0.43\textwidth]{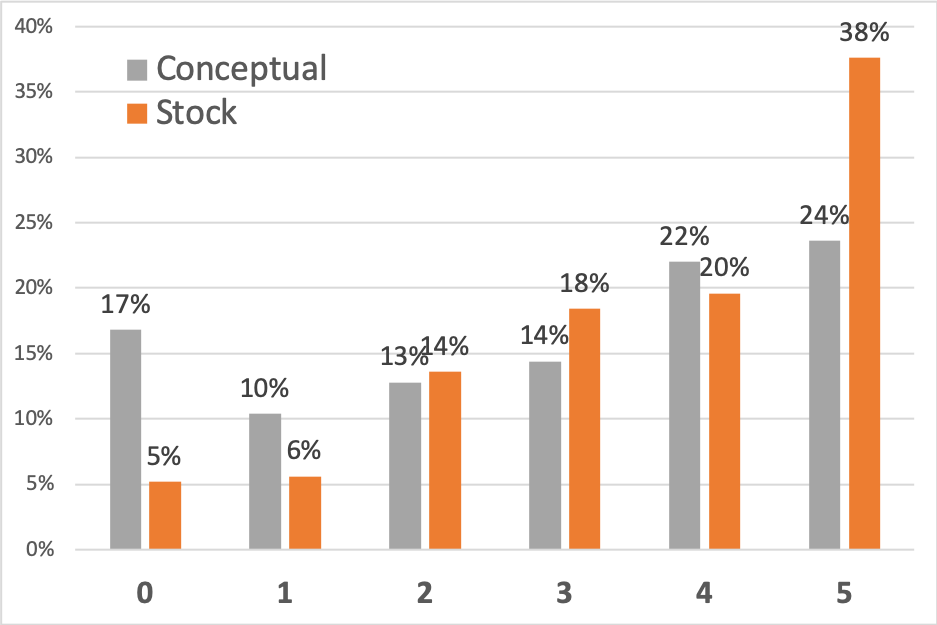}
\end{center}
   \caption{Distribution of human-given scores for our two constructed datasets. Dataset constructed on Stock Captions gets higher human scores. }
\label{fig:user}
\end{figure}

To better compare the two 
datasets, we ask the users to give strict 
scores. A caption needs to be precise, discriminative and natural
to be considered good. Many captions with the score of 0 and 1 are 
semantically good, but are unnatural.
The distribution of scores is shown in 
Figure \ref{fig:user}. As expected, in overall quality, the Stock Captions data scores 
significantly higher as it is based on compact and precise human-generated 
captions. However, several users do note that the captions in
the Conceptual Captions dataset seems to be more specific, and 
``interesting''.

\section{Method}
\begin{figure*}
\begin{center}
\includegraphics[width=1.02\linewidth]{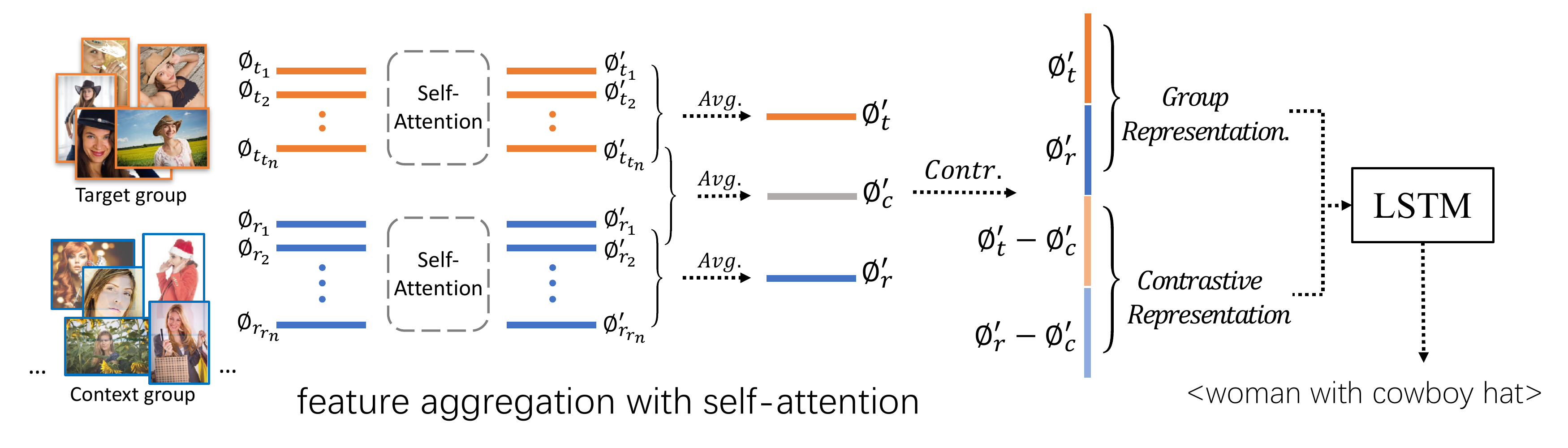}
\end{center}
   \caption{Context-aware group captioning with self-attention and contrastive features. Image features are aggregated with self-attention to get the group representation for each image group. Then the group representation is concatenated with contrastive representation to compose the input to LSTM decoder, which finally generates context-aware caption for the target image group.}
\label{fig:pipeline}
\end{figure*}


In this section, we explore methods to address the two main challenges in our proposed problem: a) \textbf{feature aggregation}, \ie how to summarize the images within one image group, and (b) \textbf{group contrasting}, \ie, how to figure out the difference between two image groups. By comparing different methods, our goal is not only finding the best performing models, but also drawing insights into the characteristics of the task, and hopefully, setting the focus for future exploration in this problem.

To begin the section, we first formalize the problem settings in 
Section~\ref{sec:prob_set}. In the subsequent sub-sections, we describe our
method explorations path starting with a simple baseline. We then gradually introduce  
more computationally specialized modules. For each module, we describe our intuition and
back them up with quantitative results and visual illustrations.

\subsection{Problem Setting}
\label{sec:prob_set}
Given a group of $n_{t}$ target images and a group of $n_{r}$ reference images, our task is to generate a description $D = <\hat{w}_{1}, ..., \hat{w}_{l}>$ to describe the target image group in context of the reference group. Here $\hat{w}_{i}$ denotes the word in the sentence and $l$ is sentence length, which varies for each data sample. In our setting, $n_{t}=5, n_{r}=15$.

Each image is represented by a 2048-d feature extracted using the ResNet50 network~\cite{he2016deep} (after pool5 layer), pretrained on ImageNet~\cite{deng2009imagenet}. 
The input of our model are the target features $\Phi_{t} = 
[\phi^{1}_{t}, ..., \phi^{n_{t}}_{t}]$ and the reference features 
$\Phi_{r} = [\phi^{1}_{r}, ..., \phi^{n_{r}}_{r}]$, where $\phi^{i} \in \mathbb{R}^{2048}$. 
We use $\Phi$ to denote a list of features, while a single feature is denoted as $\phi$.

While we believe that more detailed features (e.g. spatial features without mean-pooling, or object-level features) may improve performance, they increase the computational complexity, and by extension, the training time to an unacceptably high level in our initial testing. Thus, we simply use the mean-pooled feature vector.


\subsection{Baseline: feature averaging and concatenation}
\label{sec:baseline}

From the problem setting above, one intuitive approach would be to summarize
the target and reference features by averaging, and concatenating them to create the
final feature for description generation. The process can be formalized as follows.

We compute the target group feature $\phi'_{t}$ and the reference group feature $\phi'_{r}$
by averaging the features in each group:

\begin{equation*}
\small{
\begin{aligned}  
\phi'_{t} = \frac{1}{n_{t}}\sum_{i \in 1..n_{t}}\phi_{t_{i}} \qquad
\phi'_{r} = \frac{1}{n_{r}}\sum_{i \in 1..n_{r}}\phi_{t_{r}}
\end{aligned}
}
\end{equation*}

Following standard captioning pipeline, we then use the concatenation of the two group features as input to LSTM to predict the context-aware descriptions. We use LSTM-RNN~\cite{hochreiter1997long} to generate the caption in an auto-regressive manner. Denoting the output of the LSTM module at time step $t$ as $h_t$, we have the equations for decoding:

\begin{equation*}
\small{
\begin{aligned}
\label{eqn:decode}
h_1 &= [\phi'_{t}, \phi'_{r}] \\
h_t &= \text{LSTM}(h_{t-1},\hat{w}_{t-1}) \\
\hat{w}_{t} &\sim \text{softmax}(h_{t}).
\end{aligned}
}
\end{equation*}

Finally, we follow standard beam search process to generate the captions. This decoding architecture is used in all of our subsequent model variants.

\subsection{Feature aggregation with self attention}

While the average-pooling method used for feature aggregation above is intuitive, it treats all image
features equally. We note that many groups
of images have prominent members that encapsulate the joint information of the whole 
groups (Figure~\ref{fig:attention}). We argue that the group summarizing process could be improved
if we can identify these prominent features/images. Motivated by that observation, we propose to use the transformer 
architecture~\cite{vaswani2017attention} for this task. The transformer relies on a grid
of attention between the elements of the set to learn a better representation. Intuitively,
by learning the self-attention grid, the model can detect the prominent features as
each element in the set can ``vote'' for the importance of the other elements through the 
attention mechanism. In the subsequent analysis, we show that, in our task,
the self-attention gird indeed puts a lot more weights to the prominent images. The
core computations of our transformer-based architecture can be summarized as follows.\footnote{We only describe the core computation steps of the self-attention due to space constraint and to improve clarity. More details can be found in the original paper~\cite{vaswani2017attention}. We also release our implementation if accepted.}

The first step is calculating the 
contextualized features using self-attention mechanism. Given the
input features $\Phi$; three different sets of features: queries $Q$, keys $K$ and values $V$ are calculated using a linear transformation:

\begin{equation*}
\small{
\begin{aligned}  
Q = W^Q \Phi + b^Q \\
K = W^K \Phi + b^K \\
V = W^V \Phi + b^V 
\end{aligned}
}
\end{equation*}

Then the self-attention grid is calculated by a scaled dot product between 
$Q$ and $K$ (the scaling factor $d$ is the dimension of the vectors in $Q$ 
and $K$). The self-attention layer uses this attention grid and the value
matrix $V$ to compute its outputs.\footnote{We don't use the multi-head attention
in this work, as in our preliminary experiments, the multi-head attention provides
no performance gain compared to a single head.}

\begin{equation*}
\small{
\begin{aligned}  
\text { Attention }(Q, K, V)=\operatorname{softmax}\left(\frac{Q K^{T}}{\sqrt{d}}\right) V
\end{aligned}
}
\end{equation*}

The self-attention output is then coupled with the residual signal to create the contextualized features $\Phi'$.

\begin{equation*}
\small{
\begin{aligned}  
V' &= V+\text {Attention}(Q, K, V) \\
\Phi' &= V' + \max \left(0, V' W_{1}+b_{1}\right) W_{2}+b_{2}
\end{aligned}
}
\end{equation*}

From this point, we denote the process of transforming from the original features set 
$\Phi$ to the contextualized feature set $\Phi'$ as $\Phi' = F(\Phi)$. With this formulation, we have the 
contextualized set of features $\Phi'_{t}$ and $\Phi'_{r}$:
\begin{equation*}
\begin{aligned}  
\Phi'_{t} = F_{st}(\Phi_{t}) \qquad
\Phi'_{r} = F_{sr}(\Phi_{r})\\
\end{aligned}
\end{equation*}
We tried both sharing and not-sharing weights of $F_{st}$ and $F_{sr}$, and found that sharing weights lead to slightly better performance. This is intuitive as the task of grouping target images are not different from the task of grouping reference images, and thus, the grouping model can share the same set of weights.

In our experiments, the self-attention architecture provides
a significant boost in performance compared to the average-pooling variant.

\subsection{Group contrasting with contrastive features}

The second major challenge in our proposed problem is the image 
group contrasting. With the aforementioned self-attention mechanism,
we have good representations for the target and reference groups. The 
most intuitive way to learn the difference between the two features
is either concatenation (which is implemented in our baseline) or 
feature subtraction.

We argue that, to learn the difference between two groups of images,
we first need to capture their similarity. Our hypothesis is 
that, when we identify the similarity between all the images,
we can ``remove'' this similarity portion from the two 
features to deduce more discriminative representation. This process
is formalized as follows.

The first step is learning the common information $\phi'_{c}$ 
between all the images. We do that by applying the same 
self-attention mechanism
described above to all the images.

\begin{equation*}
\small{
\begin{aligned}
    \Phi'_{c} &= F_{a}([\Phi_{t}; \Phi_{r}]) \\
    \phi'_{c} &= \frac{1}{n_t + n_r} \sum \Phi'_{c}
\end{aligned}
}
\end{equation*}

Then the joint information is ``removed'' from the group features 
$\phi'_{t}$ and $\phi'_{r}$ by subtraction to generate the contrastive/residual feature $\phi^d_{t}$ and  $\phi^d_{r}$.

\begin{equation*}
\small{
\begin{aligned}
    \phi^d_{t} = \phi'_{t} - \phi'_{c} \qquad
    \phi^d_{r} = \phi'_{r} - \phi'_{c}
\end{aligned}
}
\end{equation*}

The contrastive features $\phi^d_{t}$ and $\phi^d_{r}$ are concatenated with the group
features $\phi'_{t}$ and $\phi'_{r}$, which are then fed into LSTM-RNN 
to generate captions. In our subsequent analysis, we show that 
the contrastive features indeed focus on the difference between two image groups.

\section{Experiments}
\newcolumntype{M}{>{\centering\arraybackslash}X}
\begin{table*}[h!]
\small{
  \begin{center}
    \begin{tabularx}{0.8\textwidth}{|l|MMMMMMM|}
        \hline
         & {WordAcc} & {CIDER} & {WER} & {BLEU1} & {BLEU2} & {METEOR} & {ROUGE}\\
        \hline
        \multicolumn{8}{|c|}{Conceptual}\\
        \hline
        Per-Img. Caption & 5.4638 & 0.4671 & 2.6587 & 0.1267 & 0.0272 & 0.0868 & 0.1466 \\
        {Average} & 36.7329 & 1.9591 & 1.6859 & 0.4932 & 0.2782 & 0.3956 & 0.4964 \\
        {SA} & 37.9916 & 2.1446 & 1.6423 & 0.5175 & 0.3103 & 0.4224 & 0.5203 \\
        {Average+Contrast} & 37.8450 & 2.0315 & 1.6534 & 0.5007 & 0.2935 & 0.4057 & 0.5027 \\
        \textbf{SA+Contrast} & \textbf{39.4496} & \textbf{2.2917} & \textbf{1.5806} & \textbf{0.5380} & \textbf{0.3313} & \textbf{0.4405} & \textbf{0.5352} \\
        \hline
        \multicolumn{8}{|c|}{Stock}\\
        \hline
        Per-Img. Caption & 5.8931 & 0.3889 & 1.8021 & 0.1445 & 0.0359 & 0.0975 & 0.1620 \\
        {Average} & 37.9428	& 1.9034 & 1.1430 & 0.5334 & 0.2429 & 0.4042 & 0.5318 \\
        {SA} & 39.2410 & 2.1023 & 1.0829 & 0.5537 & 0.2696 & 0.4243 & 0.5515 \\
        {Average+Contrast} & 39.1985 & 2.0278 & 1.0956 & 0.5397 & 0.2632 & 0.4139 & 0.5375\\
        \textbf{SA+Contrast} & \textbf{40.6113} & \textbf{2.1561} & \textbf{1.0529} & \textbf{0.5601} & \textbf{0.2796} & \textbf{0.4332} & \textbf{0.5572} \\
        \hline
        \end{tabularx}
  \end{center}
  \caption{Group captioning performance on the Conceptual Captions and Stock Captions dataset.}\vspace{-0.1in}
  \label{tab:results}
}
\end{table*}

In this section, we first describe our evaluation results on the two datasets. 
Then we provide quantitative analysis and visualization to expose the effectiveness of 
different components of our model.


\begin{figure}
\begin{center}
\includegraphics[width=0.48\textwidth]{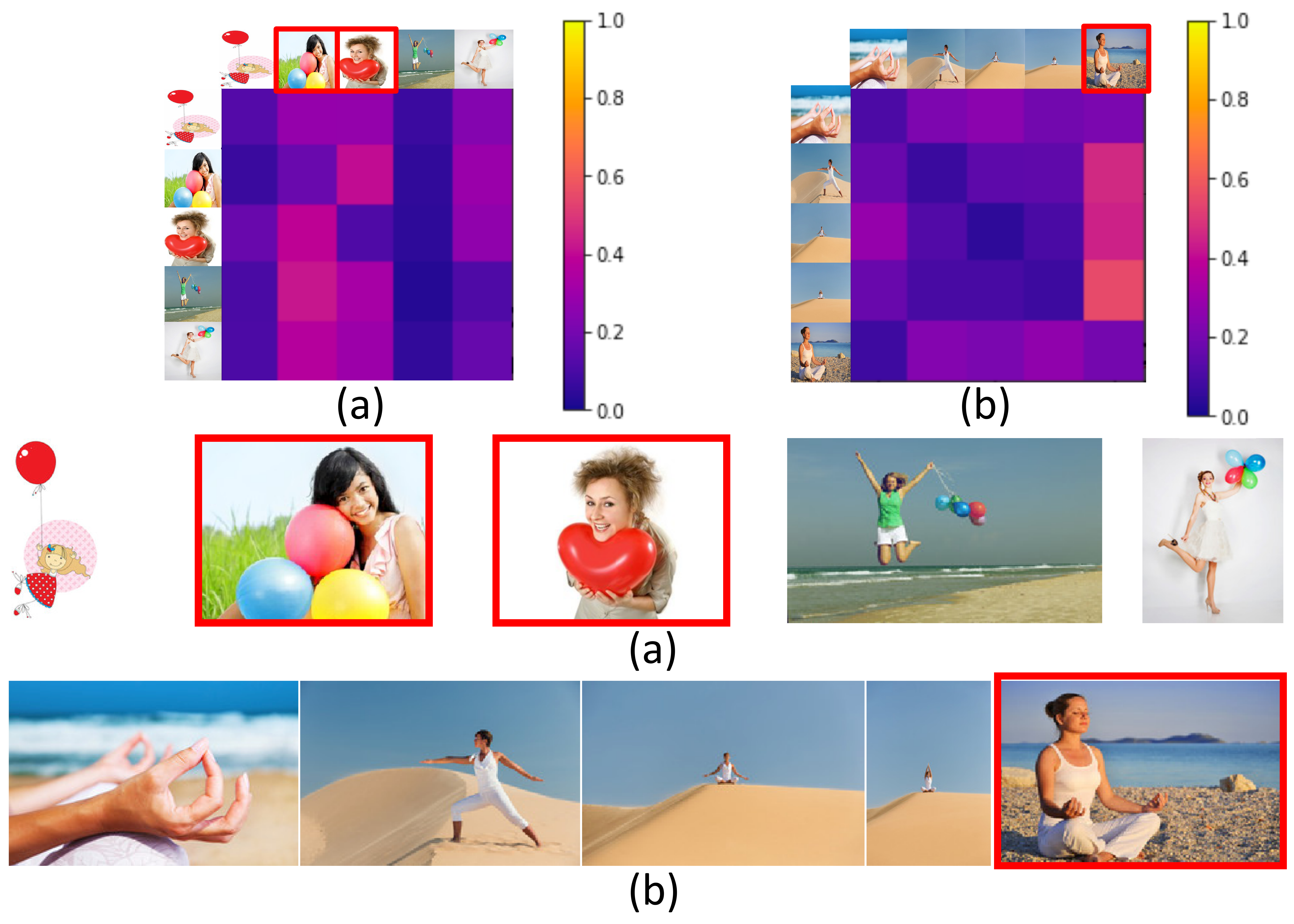}
\end{center}
   \caption{Visualization of $5\times5$ self-attention weight matrix for target image group. Each row sums up to 1. For group (a) \texttt{woman with balloon}, image 2 and image 3 are representative. For group (b) \texttt{yoga on beach}, image5 is representative. Images with more distinguishable features become the representative images of a group and get higher attention weights.}
\label{fig:attention}
\end{figure}

\setlength{\tabcolsep}{0.1cm}
\begin{table}
\footnotesize{
  \begin{center}
    \begin{tabular}{|l|r|r|r|r|r|}
        \hline
        {Model} & {WordAcc} & {CIDER} & {BLEU2} & {METEOR} & {ROUGE}\\
        \hline
        {Tgt0 + Ref15} & 24.4709 & 1.0399 & 0.0614 & 0.2341 & 0.3965 \\
        {Tgt1 + Ref15} & 28.7479 & 1.3447 & 0.1292 & 0.2938 & 0.4415 \\
        {Tgt3 + Ref15} & 34.6574 & 1.7641 & 0.2098 & 0.3698 & 0.5048 \\
        {Tgt5 + Ref0} & 31.8061 & 1.6767 & 0.2095 & 0.3475 & 0.4552 \\
        \textbf{Tgt5 + Ref15} & \textbf{40.6113} & \textbf{2.1561} & \textbf{0.2796} & \textbf{0.4332} & \textbf{0.5572} \\
    \hline
        \end{tabular}
  \end{center}
  \caption{Performance with varying the number of target and reference images. (evaluated on Stock Captions dataset)}\vspace{-0.2in}
  \label{tab:number}
  }
\end{table}

\subsection{Group Captioning Performance}

We evaluate our context-aware group captioning method on both Conceptual Captions and Stock Captions datasets. The same hyper-parameters are used for all experiments on each dataset. On the Stock Captions dataset, we use batch size 512 and initial learning rate $1\times10^{-4}$. On the Conceptual Captions dataset, we use batch size 512 and learning rate $5\times10^{-5}$. We train the model for 100 epochs with Adam optimizer\cite{kingma2014adam} on both datasets. 

We measure the captioning performance on the test splits in both datasets using a variety of captioning metrics. Specifically, we consider the standard metrics widely used in image captioning literature, including BLEU\cite{papineni2002bleu}, CIDER\cite{vedantam2015cider}, METEOR\cite{banerjee2005meteor} and ROUGE\cite{lin2004rouge}. In addition, since group descriptions are often short and compact, we put more emphasis on single word accuracy compared to traditional image captioning. We thus consider two additional metrics,  Word-by-word accuracy(WordAcc), word error rate(WER), that specifically assess word-based accuracy\footnote{Here we consider position-specific word accuracy. For example, prediction \emph{woman with straw hat} with ground truth \emph{woman with cowboy hat} has accuracy 75\%, while prediction \emph{woman with hat} has accuracy 50\%.}. 
We also note that as some group descriptions may contain as few as two words, we do not consider BLEU3 and BLEU4 scores which evaluates tri-grams and 4-grams. 


The captioning performance on the testing set of Conceptual Captions and Stock Captions datasets are reported in Table \ref{tab:results}. 
To compare with a simple baseline, we caption each image individually and summarize them using our dataset building method. The result (\textbf{Per-Img. Caption}) shows that the group captioning problem cannot be solved by simply summarizing per-image captions. More details are shown in supplementary materials.
Compared to aggregating features by averaging (\textbf{Average}, as in Section \ref{sec:baseline}), self-attention (\textbf{SA}) is more effective in computing group representation and leads to significant performance improvement. 
On top of feature aggregation, contrastive feature is critical for the model to generate context-aware caption which emphasizes the difference of target image group on context of reference group. Applying contrastive features (\textbf{Contrast}) to either feature aggregation methods leads to performance boost (\textbf{Average+Contrast, SA+Contrast}). To this end, our full model, which combines self-attention for group aggregation and contrastive feature for group comparing performs best, achieving 39.4\% WordAcc on Conceptual Captions and 40.6\% on Stock Captions. 

\subsection{Discussion}

\begin{figure*}[h!]\vspace{-0.05in}
\begin{center}
\includegraphics[width=0.99\linewidth]{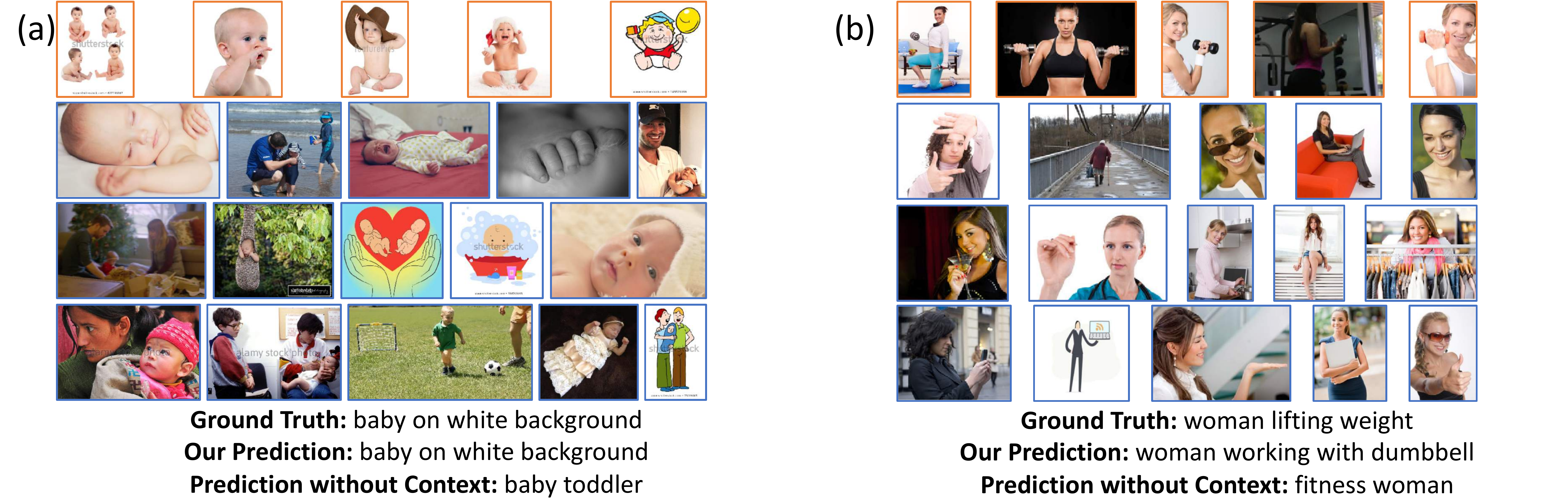}
\end{center}
   \caption{Qualitative prediction examples on Conceptual Captions (a) and Stock Captions (b) datasets. In each example, images in first row (in orange boxes) are target images while second to fourth rows (in blue boxes) are reference images. Our model can effectively summarize relevant information in the image groups during captioning. Our model also effectively takes discriminative information between the target and reference group into account during captioning to predict accurate group captioning results.}\vspace{-0.1in}
\label{fig:example}
\end{figure*}

\noindent
\textbf{Effectiveness of self-attention on feature aggregation.}
To better understand the effectiveness of self-attention, in Figure \ref{fig:attention}, we visualize the $5\times5$ self-attention weight matrix between 5 target images.
The i-th row of the attention matrix represents the attention weights from i-th image to each of the 5 images, which sum up to 1. 
In (a), images with larger and centered balloons (Image2 and Image3) gets higher attention.
In (b), image5 where the woman doing yoga is larger and easier to recognize gets higher attention.
In both examples, images with more recognizable features get higher attention weights and thus contribute more to the aggregated group representation.

\begin{table}[h!]
\small{
  \begin{center}
    \begin{tabularx}{1.01\columnwidth}{|p{3.05cm}|p{0.88cm}|X|}
        \hline
        \multicolumn{1}{|c|}{\textbf{Contrastive + Group}} & \multicolumn{1}{c}{\textbf{Group}} & \multicolumn{1}{|c|}{\textbf{Contrastive}}\\
        \hline
        \textcolor{blue}{woman} with \textcolor{red}{cowboy hat} & \textcolor{blue}{woman} & country with \textcolor{red}{cowboy} straw \textcolor{red}{hat}\\
        \hline
        \textcolor{red}{white} \textcolor{blue}{girl} & \textcolor{blue}{girl} & \textcolor{red}{white} rule \textcolor{red}{white} and...\\
        \hline
        \textcolor{blue}{woman} in \textcolor{red}{boxing glove} & \textcolor{blue}{woman} & is go in \textcolor{red}{boxing}... \\
        \hline
        \end{tabularx}
  \end{center}
  \caption{Analysis of contrastive representation. Column \texttt{Contrastive + Group} is the prediction of our full model. Column \texttt{Group} and column \texttt{Contrastive} are the predictions when only the group or only the contrastive representation is fed into the decoder respectively. \textcolor{blue}{Blue} text denotes the common part while \textcolor{red}{red} text denotes the contrastive part.}\vspace{-0.25in}
  \label{tab:diff}
  }
\end{table}

\noindent
\textbf{Importance of multiple target and reference images.}
To investigate the effectiveness of giving multiple images in each group, we vary the number of target and reference images. 
Results are shown in Table \ref{tab:number}. 
Fewer target or reference images results in performance decline, which indicates that a larger number of images is more informational for the model to get better descriptions. 
We also qualitatively study the importance of the reference image group. Examples are shown in Figure \ref{fig:example}. The examples indicate that when not giving reference group the predictions tend to be more generic and less discriminative.


\noindent
\textbf{Contrastive representation versus group representation.}
Table \ref{tab:diff} shows example descriptions when only the group representations or only the contrastive representations are fed into LSTM decoder. 
Although the model does not treat the features
independently and removing the features might 
break the grammar structure of the caption, 
looking at the lexicons returned by the two 
variants, we can clearly observe the focus of
two features. When the decoder uses only the group representations, the predictions emphasize the common part of two image groups. On the other hand, when the decoder only uses the contrastive representations, the predictions emphasize the difference between two image groups.
This reveals that the group representation encodes similarity information, while the contrastive representation encodes discriminative information.

\begin{figure}[h!]\vspace{-0.05in}
\begin{center}
\includegraphics[width=0.99\linewidth]{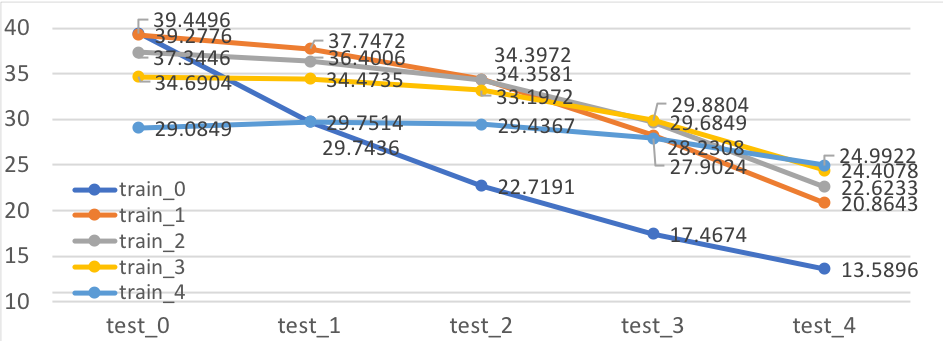}
\end{center}
   \caption{Performance change on Conceptual Captions dataset when trained and tested with 0-4 random images in the target group. Training with more noise increases robustness of the model but hinder performance when tested with no noise.}\vspace{-0.25in}
\label{fig:noise}
\end{figure}

\noindent
\textbf{Robustness to noise images.} 
To investigate the model's robustness to noise in the image group, we tried adding random unrelated images to the target group. Figure \ref{fig:noise} shows performances of models trained and tested with different number (0-4) of noise images on Conceptual Captions dataset. Training with more noise increases robustness of the model but hinder performance when tested with no noise. The model shows robustness to small noise. Qualitatively, when testing with small (1 or 2) noise (trained with 0 noise), the caption loses details, e.g. woman in red dress becomes woman in dress. The generated caption is broken when the noise is severe, which is reasonable.

\section{Conclusion}

In this paper, we present the novel context-aware group captioning task, where the objective is to describe a target image group in contrast to a reference image group. To explore this problem, we introduce two
large scale datasets, Conceptual Captions and Stock Captions respectively, both of which will be released for future research.
We also propose a framework with self-attention for grouping the images and contrastive representation for capturing discriminative features.
We show the effectiveness of our proposed model both quantitatively and qualitatively on our datasets. We also thoroughly analyze the behavior of our models to provide insights into this new problem.




\newpage
{\small
\bibliographystyle{plain}
\bibliography{egbib}

\begin{thebibliography}{10}

\bibitem{anderson2018bottom}
Peter Anderson, Xiaodong He, Chris Buehler, Damien Teney, Mark Johnson, Stephen
  Gould, and Lei Zhang.
\newblock Bottom-up and top-down attention for image captioning and visual
  question answering.
\newblock In {\em Proceedings of the IEEE Conference on Computer Vision and
  Pattern Recognition}, pages 6077--6086, 2018.

\bibitem{andreas2016reasoning}
Jacob Andreas and Dan Klein.
\newblock Reasoning about pragmatics with neural listeners and speakers.
\newblock {\em arXiv preprint arXiv:1604.00562}, 2016.

\bibitem{bai2018survey}
Shuang Bai and Shan An.
\newblock A survey on automatic image caption generation.
\newblock {\em Neurocomputing}, 311:291--304, 2018.

\bibitem{banerjee2005meteor}
Satanjeev Banerjee and Alon Lavie.
\newblock Meteor: An automatic metric for mt evaluation with improved
  correlation with human judgments.
\newblock In {\em Proceedings of the acl workshop on intrinsic and extrinsic
  evaluation measures for machine translation and/or summarization}, pages
  65--72, 2005.

\bibitem{cao2019gcnet}
Yue Cao, Jiarui Xu, Stephen Lin, Fangyun Wei, and Han Hu.
\newblock Gcnet: Non-local networks meet squeeze-excitation networks and
  beyond.
\newblock In {\em Proceedings of the IEEE International Conference on Computer
  Vision Workshops}, pages 0--0, 2019.

\bibitem{chen2018groupcap}
Fuhai Chen, Rongrong Ji, Xiaoshuai Sun, Yongjian Wu, and Jinsong Su.
\newblock Groupcap: Group-based image captioning with structured relevance and
  diversity constraints.
\newblock In {\em Proceedings of the IEEE Conference on Computer Vision and
  Pattern Recognition}, pages 1345--1353, 2018.

\bibitem{chen2017sca}
Long Chen, Hanwang Zhang, Jun Xiao, Liqiang Nie, Jian Shao, Wei Liu, and
  Tat-Seng Chua.
\newblock Sca-cnn: Spatial and channel-wise attention in convolutional networks
  for image captioning.
\newblock In {\em Proceedings of the IEEE conference on computer vision and
  pattern recognition}, pages 5659--5667, 2017.

\bibitem{dai2017towards}
Bo~Dai, Sanja Fidler, Raquel Urtasun, and Dahua Lin.
\newblock Towards diverse and natural image descriptions via a conditional gan.
\newblock In {\em Proceedings of the IEEE International Conference on Computer
  Vision}, pages 2970--2979, 2017.

\bibitem{dai2017contrastive}
Bo~Dai and Dahua Lin.
\newblock Contrastive learning for image captioning.
\newblock In {\em Advances in Neural Information Processing Systems}, pages
  898--907, 2017.

\bibitem{dehghani2017learning}
Mostafa Dehghani, Sascha Rothe, Enrique Alfonseca, and Pascal Fleury.
\newblock Learning to attend, copy, and generate for session-based query
  suggestion.
\newblock In {\em Proceedings of the 2017 ACM on Conference on Information and
  Knowledge Management}, pages 1747--1756. ACM, 2017.

\bibitem{deng2009imagenet}
Jia Deng, Wei Dong, Richard Socher, Li-Jia Li, Kai Li, and Li~Fei-Fei.
\newblock Imagenet: A large-scale hierarchical image database.
\newblock In {\em 2009 IEEE conference on computer vision and pattern
  recognition}, pages 248--255. Ieee, 2009.

\bibitem{devlin2018bert}
Jacob Devlin, Ming-Wei Chang, Kenton Lee, and Kristina Toutanova.
\newblock Bert: Pre-training of deep bidirectional transformers for language
  understanding.
\newblock {\em arXiv preprint arXiv:1810.04805}, 2018.

\bibitem{donahue2015long}
Jeffrey Donahue, Lisa Anne~Hendricks, Sergio Guadarrama, Marcus Rohrbach,
  Subhashini Venugopalan, Kate Saenko, and Trevor Darrell.
\newblock Long-term recurrent convolutional networks for visual recognition and
  description.
\newblock In {\em Proceedings of the IEEE conference on computer vision and
  pattern recognition}, pages 2625--2634, 2015.

\bibitem{he2016deep}
Kaiming He, Xiangyu Zhang, Shaoqing Ren, and Jian Sun.
\newblock Deep residual learning for image recognition.
\newblock In {\em Proceedings of the IEEE conference on computer vision and
  pattern recognition}, pages 770--778, 2016.

\bibitem{hochreiter1997long}
Sepp Hochreiter and J{\"u}rgen Schmidhuber.
\newblock Long short-term memory.
\newblock {\em Neural computation}, 9(8):1735--1780, 1997.

\bibitem{hossain2019comprehensive}
MD~Hossain, Ferdous Sohel, Mohd~Fairuz Shiratuddin, and Hamid Laga.
\newblock A comprehensive survey of deep learning for image captioning.
\newblock {\em ACM Computing Surveys (CSUR)}, 51(6):118, 2019.

\bibitem{huang2016visual}
Ting-Hao Huang, Francis Ferraro, Nasrin Mostafazadeh, Ishan Misra, Aishwarya
  Agrawal, Jacob Devlin, Ross Girshick, Xiaodong He, Pushmeet Kohli, Dhruv
  Batra, et~al.
\newblock Visual storytelling.
\newblock In {\em Proceedings of the 2016 Conference of the North American
  Chapter of the Association for Computational Linguistics: Human Language
  Technologies}, pages 1233--1239, 2016.

\bibitem{jiang2018rin}
Jyun-Yu Jiang and Wei Wang.
\newblock Rin: Reformulation inference network for context-aware query
  suggestion.
\newblock In {\em Proceedings of the 27th ACM International Conference on
  Information and Knowledge Management}, pages 197--206. ACM, 2018.

\bibitem{jiang2018recurrent}
Wenhao Jiang, Lin Ma, Yu-Gang Jiang, Wei Liu, and Tong Zhang.
\newblock Recurrent fusion network for image captioning.
\newblock In {\em Proceedings of the European Conference on Computer Vision
  (ECCV)}, pages 499--515, 2018.

\bibitem{johnson2016densecap}
Justin Johnson, Andrej Karpathy, and Li~Fei-Fei.
\newblock Densecap: Fully convolutional localization networks for dense
  captioning.
\newblock In {\em Proceedings of the IEEE Conference on Computer Vision and
  Pattern Recognition}, pages 4565--4574, 2016.

\bibitem{karpathy2015deep}
Andrej Karpathy and Li~Fei-Fei.
\newblock Deep visual-semantic alignments for generating image descriptions.
\newblock In {\em Proceedings of the IEEE conference on computer vision and
  pattern recognition}, pages 3128--3137, 2015.

\bibitem{kazemzadeh2014referitgame}
Sahar Kazemzadeh, Vicente Ordonez, Mark Matten, and Tamara Berg.
\newblock Referitgame: Referring to objects in photographs of natural scenes.
\newblock In {\em Proceedings of the 2014 conference on empirical methods in
  natural language processing (EMNLP)}, pages 787--798, 2014.

\bibitem{kim2019dense}
Dong-Jin Kim, Jinsoo Choi, Tae-Hyun Oh, and In~So Kweon.
\newblock Dense relational captioning: Triple-stream networks for
  relationship-based captioning.
\newblock In {\em Proceedings of the IEEE Conference on Computer Vision and
  Pattern Recognition}, pages 6271--6280, 2019.

\bibitem{kingma2014adam}
Diederik~P Kingma and Jimmy Ba.
\newblock Adam: A method for stochastic optimization.
\newblock {\em arXiv preprint arXiv:1412.6980}, 2014.

\bibitem{krishna2017visual}
Ranjay Krishna, Yuke Zhu, Oliver Groth, Justin Johnson, Kenji Hata, Joshua
  Kravitz, Stephanie Chen, Yannis Kalantidis, Li-Jia Li, David~A Shamma, et~al.
\newblock Visual genome: Connecting language and vision using crowdsourced
  dense image annotations.
\newblock {\em International Journal of Computer Vision}, 123(1):32--73, 2017.

\bibitem{krizhevsky2012imagenet}
Alex Krizhevsky, Ilya Sutskever, and Geoffrey~E Hinton.
\newblock Imagenet classification with deep convolutional neural networks.
\newblock In {\em Advances in neural information processing systems}, pages
  1097--1105, 2012.

\bibitem{li2019visualbert}
Liunian~Harold Li, Mark Yatskar, Da~Yin, Cho-Jui Hsieh, and Kai-Wei Chang.
\newblock Visualbert: A simple and performant baseline for vision and language.
\newblock {\em arXiv preprint arXiv:1908.03557}, 2019.

\bibitem{li2020neural}
Yingwei Li, Xiaojie Jin, Jieru Mei, Xiaochen Lian, Linjie Yang, Cihang Xie,
  Qihang Yu, Yuyin Zhou, Song Bai, and Alan Yuille.
\newblock Neural architecture search for lightweight non-local networks.
\newblock In {\em CVPR}, 2020.

\bibitem{lin2004rouge}
Chin-Yew Lin.
\newblock Rouge: A package for automatic evaluation of summaries.
\newblock In {\em Text summarization branches out}, pages 74--81, 2004.

\bibitem{liu2017attention}
Chenxi Liu, Junhua Mao, Fei Sha, and Alan Yuille.
\newblock Attention correctness in neural image captioning.
\newblock In {\em Thirty-First AAAI Conference on Artificial Intelligence},
  2017.

\bibitem{liu2017improved}
Siqi Liu, Zhenhai Zhu, Ning Ye, Sergio Guadarrama, and Kevin Murphy.
\newblock Improved image captioning via policy gradient optimization of spider.
\newblock In {\em Proceedings of the IEEE international conference on computer
  vision}, pages 873--881, 2017.

\bibitem{liu2019survey}
Xiaoxiao Liu, Qingyang Xu, and Ning Wang.
\newblock A survey on deep neural network-based image captioning.
\newblock {\em The Visual Computer}, 35(3):445--470, 2019.

\bibitem{liu2018show}
Xihui Liu, Hongsheng Li, Jing Shao, Dapeng Chen, and Xiaogang Wang.
\newblock Show, tell and discriminate: Image captioning by self-retrieval with
  partially labeled data.
\newblock In {\em Proceedings of the European Conference on Computer Vision
  (ECCV)}, pages 338--354, 2018.

\bibitem{lu2019vilbert}
Jiasen Lu, Dhruv Batra, Devi Parikh, and Stefan Lee.
\newblock Vilbert: Pretraining task-agnostic visiolinguistic representations
  for vision-and-language tasks.
\newblock {\em arXiv preprint arXiv:1908.02265}, 2019.

\bibitem{lu2017knowing}
Jiasen Lu, Caiming Xiong, Devi Parikh, and Richard Socher.
\newblock Knowing when to look: Adaptive attention via a visual sentinel for
  image captioning.
\newblock In {\em Proceedings of the IEEE conference on computer vision and
  pattern recognition}, pages 375--383, 2017.

\bibitem{luo2018discriminability}
Ruotian Luo, Brian Price, Scott Cohen, and Gregory Shakhnarovich.
\newblock Discriminability objective for training descriptive captions.
\newblock In {\em Proceedings of the IEEE Conference on Computer Vision and
  Pattern Recognition}, pages 6964--6974, 2018.

\bibitem{luo2017comprehension}
Ruotian Luo and Gregory Shakhnarovich.
\newblock Comprehension-guided referring expressions.
\newblock In {\em Proceedings of the IEEE Conference on Computer Vision and
  Pattern Recognition}, pages 7102--7111, 2017.

\bibitem{mao2016generation}
Junhua Mao, Jonathan Huang, Alexander Toshev, Oana Camburu, Alan~L Yuille, and
  Kevin Murphy.
\newblock Generation and comprehension of unambiguous object descriptions.
\newblock In {\em Proceedings of the IEEE conference on computer vision and
  pattern recognition}, pages 11--20, 2016.

\bibitem{ordonez2011im2text}
Vicente Ordonez, Girish Kulkarni, and Tamara~L Berg.
\newblock Im2text: Describing images using 1 million captioned photographs.
\newblock In {\em Advances in neural information processing systems}, pages
  1143--1151, 2011.

\bibitem{papineni2002bleu}
Kishore Papineni, Salim Roukos, Todd Ward, and Wei-Jing Zhu.
\newblock Bleu: a method for automatic evaluation of machine translation.
\newblock In {\em Proceedings of the 40th annual meeting on association for
  computational linguistics}, pages 311--318. Association for Computational
  Linguistics, 2002.

\bibitem{park2019robust}
Dong~Huk Park, Trevor Darrell, and Anna Rohrbach.
\newblock Robust change captioning.
\newblock In {\em Proceedings of the IEEE International Conference on Computer
  Vision}, pages 4624--4633, 2019.

\bibitem{rennie2017self}
Steven~J Rennie, Etienne Marcheret, Youssef Mroueh, Jerret Ross, and Vaibhava
  Goel.
\newblock Self-critical sequence training for image captioning.
\newblock In {\em Proceedings of the IEEE Conference on Computer Vision and
  Pattern Recognition}, pages 7008--7024, 2017.

\bibitem{sharma2018conceptual}
Piyush Sharma, Nan Ding, Sebastian Goodman, and Radu Soricut.
\newblock Conceptual captions: A cleaned, hypernymed, image alt-text dataset
  for automatic image captioning.
\newblock In {\em Proceedings of the 56th Annual Meeting of the Association for
  Computational Linguistics (Volume 1: Long Papers)}, pages 2556--2565, 2018.

\bibitem{shetty2017speaking}
Rakshith Shetty, Marcus Rohrbach, Lisa Anne~Hendricks, Mario Fritz, and Bernt
  Schiele.
\newblock Speaking the same language: Matching machine to human captions by
  adversarial training.
\newblock In {\em Proceedings of the IEEE International Conference on Computer
  Vision}, pages 4135--4144, 2017.

\bibitem{sordoni2015hierarchical}
Alessandro Sordoni, Yoshua Bengio, Hossein Vahabi, Christina Lioma, Jakob
  Grue~Simonsen, and Jian-Yun Nie.
\newblock A hierarchical recurrent encoder-decoder for generative context-aware
  query suggestion.
\newblock In {\em Proceedings of the 24th ACM International on Conference on
  Information and Knowledge Management}, pages 553--562. ACM, 2015.

\bibitem{suhr2017corpus}
Alane Suhr, Mike Lewis, James Yeh, and Yoav Artzi.
\newblock A corpus of natural language for visual reasoning.
\newblock In {\em Proceedings of the 55th Annual Meeting of the Association for
  Computational Linguistics (Volume 2: Short Papers)}, pages 217--223, 2017.

\bibitem{suhr2018corpus}
Alane Suhr, Stephanie Zhou, Ally Zhang, Iris Zhang, Huajun Bai, and Yoav Artzi.
\newblock A corpus for reasoning about natural language grounded in
  photographs.
\newblock {\em arXiv preprint arXiv:1811.00491}, 2018.

\bibitem{sun2019videobert}
Chen Sun, Austin Myers, Carl Vondrick, Kevin Murphy, and Cordelia Schmid.
\newblock Videobert: A joint model for video and language representation
  learning.
\newblock {\em arXiv preprint arXiv:1904.01766}, 2019.

\bibitem{tan2019expressing}
Hao Tan, Franck Dernoncourt, Zhe Lin, Trung Bui, and Mohit Bansal.
\newblock Expressing visual relationships via language.
\newblock {\em arXiv preprint arXiv:1906.07689}, 2019.

\bibitem{vaswani2017attention}
Ashish Vaswani, Noam Shazeer, Niki Parmar, Jakob Uszkoreit, Llion Jones,
  Aidan~N Gomez, {\L}ukasz Kaiser, and Illia Polosukhin.
\newblock Attention is all you need.
\newblock In {\em Advances in neural information processing systems}, pages
  5998--6008, 2017.

\bibitem{vedantam2017context}
Ramakrishna Vedantam, Samy Bengio, Kevin Murphy, Devi Parikh, and Gal Chechik.
\newblock Context-aware captions from context-agnostic supervision.
\newblock In {\em Proceedings of the IEEE Conference on Computer Vision and
  Pattern Recognition}, pages 251--260, 2017.

\bibitem{vedantam2015cider}
Ramakrishna Vedantam, C~Lawrence~Zitnick, and Devi Parikh.
\newblock Cider: Consensus-based image description evaluation.
\newblock In {\em Proceedings of the IEEE conference on computer vision and
  pattern recognition}, pages 4566--4575, 2015.

\bibitem{vinyals2015show}
Oriol Vinyals, Alexander Toshev, Samy Bengio, and Dumitru Erhan.
\newblock Show and tell: A neural image caption generator.
\newblock In {\em Proceedings of the IEEE conference on computer vision and
  pattern recognition}, pages 3156--3164, 2015.

\bibitem{wang2016image}
Cheng Wang, Haojin Yang, Christian Bartz, and Christoph Meinel.
\newblock Image captioning with deep bidirectional lstms.
\newblock In {\em Proceedings of the 24th ACM international conference on
  Multimedia}, pages 988--997. ACM, 2016.

\bibitem{wang2018non}
Xiaolong Wang, Ross Girshick, Abhinav Gupta, and Kaiming He.
\newblock Non-local neural networks.
\newblock In {\em Proceedings of the IEEE conference on computer vision and
  pattern recognition}, pages 7794--7803, 2018.

\bibitem{wang2018scene}
Yu-Siang Wang, Chenxi Liu, Xiaohui Zeng, and Alan Yuille.
\newblock Scene graph parsing as dependency parsing.
\newblock {\em arXiv preprint arXiv:1803.09189}, 2018.

\bibitem{wu2018query}
Bin Wu, Chenyan Xiong, Maosong Sun, and Zhiyuan Liu.
\newblock Query suggestion with feedback memory network.
\newblock In {\em Proceedings of the 2018 World Wide Web Conference}, pages
  1563--1571. International World Wide Web Conferences Steering Committee,
  2018.

\bibitem{xu2015show}
Kelvin Xu, Jimmy Ba, Ryan Kiros, Kyunghyun Cho, Aaron Courville, Ruslan
  Salakhudinov, Rich Zemel, and Yoshua Bengio.
\newblock Show, attend and tell: Neural image caption generation with visual
  attention.
\newblock In {\em International conference on machine learning}, pages
  2048--2057, 2015.

\bibitem{yang2017dense}
Linjie Yang, Kevin Tang, Jianchao Yang, and Li-Jia Li.
\newblock Dense captioning with joint inference and visual context.
\newblock In {\em Proceedings of the IEEE Conference on Computer Vision and
  Pattern Recognition}, pages 2193--2202, 2017.

\bibitem{yao2018exploring}
Ting Yao, Yingwei Pan, Yehao Li, and Tao Mei.
\newblock Exploring visual relationship for image captioning.
\newblock In {\em Proceedings of the European Conference on Computer Vision
  (ECCV)}, pages 684--699, 2018.

\bibitem{you2016image}
Quanzeng You, Hailin Jin, Zhaowen Wang, Chen Fang, and Jiebo Luo.
\newblock Image captioning with semantic attention.
\newblock In {\em Proceedings of the IEEE conference on computer vision and
  pattern recognition}, pages 4651--4659, 2016.

\bibitem{yu2016modeling}
Licheng Yu, Patrick Poirson, Shan Yang, Alexander~C Berg, and Tamara~L Berg.
\newblock Modeling context in referring expressions.
\newblock In {\em European Conference on Computer Vision}, pages 69--85.
  Springer, 2016.

\bibitem{yu2017joint}
Licheng Yu, Hao Tan, Mohit Bansal, and Tamara~L Berg.
\newblock A joint speaker-listener-reinforcer model for referring expressions.
\newblock In {\em Proceedings of the IEEE Conference on Computer Vision and
  Pattern Recognition}, pages 7282--7290, 2017.

\bibitem{yue2018compact}
Kaiyu Yue, Ming Sun, Yuchen Yuan, Feng Zhou, Errui Ding, and Fuxin Xu.
\newblock Compact generalized non-local network.
\newblock In {\em Advances in Neural Information Processing Systems}, pages
  6510--6519, 2018.

\bibitem{zha2009visual}
Zheng-Jun Zha, Linjun Yang, Tao Mei, Meng Wang, and Zengfu Wang.
\newblock Visual query suggestion.
\newblock In {\em Proceedings of the 17th ACM international conference on
  Multimedia}, pages 15--24. ACM, 2009.

\bibitem{zha2010_TOMM}
Zheng-Jun Zha, Linjun Yang, Tao Mei, Meng Wang, Zengfu Wang, Tat-Seng Chua, and
  Xian-Sheng Hua.
\newblock Visual query suggestion: Towards capturing user intent in internet
  image search.
\newblock {\em ACM Trans. Multimedia Comput. Commun. Appl.}, 6(3), August 2010.

\bibitem{zhu2019asymmetric}
Zhen Zhu, Mengde Xu, Song Bai, Tengteng Huang, and Xiang Bai.
\newblock Asymmetric non-local neural networks for semantic segmentation.
\newblock In {\em Proceedings of the IEEE International Conference on Computer
  Vision}, pages 593--602, 2019.

\end{thebibliography}
}

\clearpage
\onecolumn
\appendix

\section{Details of Datasets}

There are six types of captions: subject-relation-object, adjective-object, noun-object, attribute-object-relation-object, object-relation-attribute-object, attribute-object-relation-attribute-object. Table \ref{tab:dataset} shows the number of samples in each type of captions.

\begin{table}[h!]
\small{
  \begin{center}
    \begin{tabular}{L{4cm} L{2cm} L{2cm}}
        \toprule
         & {Conceptual} & {Stock}\\
        \hline
        Sub-Rel-Obj & 46810 & 40620 \\
        Adj-Obj & 24890 & 33650 \\
        NN-Obj & 18466 & 32774 \\
        Att-Sub-Rel-Obj & 55124 & 19170 \\
        Sub-Rel-Att-Obj & 30944 & 16683 \\
        Att-Sub-Rel-Att-Obj & 23208 & 3442 \\
        \hline
        Total & 199442 & 146339 \\
        \bottomrule
    \end{tabular}
  \end{center}
  \caption{Statistics of each caption type on Conceptual Captions and Stock Captions.}
  \label{tab:dataset}
 }
 \end{table}

\section{Experiments}

\subsection{Varying the Number of Reference Images}

In Table 3 of the main paper, we give experiment results of varying the number of target and reference images. Here in Table \ref{tab:number} we give more detailed results of varying the number of reference images. As shown in the table, the performance improves when more reference images are given. We also notice that while the differences between giving 0, 5 or 10 references images are large, the gap between 10 and 15 reference images are not significant. So we use 15 reference images in the overall experiment setting.

\newcolumntype{M}{>{\centering\arraybackslash}X}
\begin{table*}[h!]
\small{
  \begin{center}
    \begin{tabularx}{0.8\textwidth}{|l|MMMMMMM|}
        \hline
         & {WordAcc} & {CIDER} & {WER} & {BLEU1} & {BLEU2} & {METEOR} & {ROUGE}\\
        \hline
        {Tgt5 + Ref0} & 31.8061 & 1.6767 & 1.2539 & 0.4600 & 0.2095 & 0.3475 & 0.4552 \\
        {Tgt5 + Ref5} & 37.1283 & 1.9536 & 1.1413 & 0.5219 & 0.2503 & 0.3987 & 0.5185 \\
        {Tgt5 + Ref10} & 39.4072 & 2.076 & 1.0923 & 0.5451 & 0.2684 & 0.4201 & 0.5424 \\
        {Tgt5 + Ref15} & 40.6113 & 2.1561 & 1.0529 & 0.5601 & 0.2796 & 0.4332 & 0.5572 \\
        \hline
        \end{tabularx}
  \end{center}
  \caption{Performance change when varying the number of reference images on Stock Captions dataset.}
  \label{tab:number}
}
\end{table*}
\subsection{Variations of Contrastive Representation}
In this subsection we show the experimental results of model variations we tried for contrasting the two image groups. The results of variation models are shown in Table \ref{tab:results}.
\subsubsection{The cross-attention models}
Given the effectiveness of attention on grouping images, we tried applying attention to contrast two image groups. We investigate three different variants:

\textbf{AttenAll}: Applying self-attention between all the target and reference images simultaneously (we use two different fully-connected layers to differentiate target and reference). This variant decreases the performance over self-attention only. We hypothesize that treating two distinct relations : intra-group relations (which the model must focus on the similarity) and inter-group relations (which the model must focus on the difference) might not be the ideal solution. Thus, we develop the second variant, which treat these two relations separately: Cross attention (\textbf{CA})

\textbf{CA}: In \textbf{CA}, we tried applying self-attention within each image group first and then cross-attention between two image groups. When doing cross-attention, we apply a mask to the self-attention kernel to remove attention connections within each image group and only keep connections between groups. This leads to slight improvement over \textbf{AttenAll}, but the performance is still behind the Self-attention only variant.

\textbf{NCA}: Going a step further, we experiment with the negative cross attention mechanism (\textbf{NCA}), which is to negate the reference image features before computing attention. The intuition is, by negating one group of features, two feature vectors that are close in the feature space will become distant. Thus, we want to force the to focus on the difference between the features, instead of the similarities. Negative cross attention improves the performance over \textbf{CA} but does not lead to consistent improvement of self-attention only. 

From the experimental results, we hypothesize that the self-attention kernel is only effective in similarity detection, not in extracting the difference, even with the negative trick. However, if we consider two feature groups as two mathematical sets, and if we can detect the common elements between the two sets, we can just ``remove’’ them from both sets and get the ``difference’’ of the two sets. This intuition leads us to the development of the contrastive representation models. Our formulation in the main paper is the translation of this intuition in neural network language.

\subsubsection{Variants of the contrastive representation model}
We also tried different variants of contrastive representation. In the method part, we derive the contrastive representation by concatenating the difference of target and reference features with their joint information, i.e., $\phi^{d} = [\phi^{d}_{t}; \phi^{d}_{r}] = [\phi'_{t} - \phi'_{c}; \phi'_{r} - \phi'_{c}]$). Besides this variant, we also tried computing contrastive representation by taking difference of target and reference features, i.e., $\phi^{d} = \phi'_{t} - \phi'_{r}$ (\textbf{SA+Contrast1}) or taking difference between target features and joint features, i.e., $\phi^{d} = \phi^{d}_{t} = \phi'_{t} - \phi'_{c}$ (\textbf{SA+Contrast2}). Both methods improve performance over self-attention (\textbf{SA}) but the results are 
lower than our best method (\textbf{SA+Contrast}), which indicates the contribution of term $\phi^{d}_{r}$ and the advantage to minus the joint information of all images instead of minus reference features.

\newcolumntype{M}{>{\centering\arraybackslash}X}
\begin{table*}[h!]
\small{
  \begin{center}
    \begin{tabularx}{0.8\textwidth}{|l|MMMMMMM|}
        \hline
         & {WordAcc} & {CIDER} & {WER} & {BLEU1} & {BLEU2} & {METEOR} & {ROUGE}\\
        \hline
        \multicolumn{8}{|c|}{Conceptual}\\
        \hline
        {Average} & 36.7329 & 1.9591 & 1.6859 & 0.4932 & 0.2782 & 0.3956 & 0.4964 \\
        {SA} & 37.9916 & 2.1446 & 1.6423 & 0.5175 & 0.3103 & 0.4224 & 0.5203 \\
        {Average+Contrast} & 37.8450 & 2.0315 & 1.6534 & 0.5007 & 0.2935 & 0.4057 & 0.5027 \\
        \textbf{SA+Contrast} & \textbf{39.4496} & \textbf{2.2917} & \textbf{1.5806} & \textbf{0.5380} & \textbf{0.3313} & \textbf{0.4405} & \textbf{0.5352} \\
        \hline
        {AttenAll} & 36.1231 & 2.0727 & 1.6851 & 0.5044 & 0.2976 & 0.4089 & 0.5059 \\
        {SA+CA} & 36.2892 & 2.1282 & 1.6697 & 0.5041 & 0.3094 & 0.4145 & 0.5062 \\
        {SA+NCA} & 37.6046 & 2.2109 & 1.6344 & 0.5155 & 0.3183 & 0.4237 & 0.5165 \\
        \hline
        {SA+Contrast1} & 38.2574 & 2.1499 & 1.6332 & 0.5213 & 0.3106 & 0.4228 & 0.5203 \\
        {SA+Contrast2} & 38.5916 & 2.1821 & 1.6230 & 0.5218 & 0.3156 & 0.4261 & 0.5229 \\
        \hline
        \multicolumn{8}{|c|}{Stock}\\
        \hline
        {Average} & 37.9428	& 1.9034 & 1.1430 & 0.5334 & 0.2429 & 0.4042 & 0.5318 \\
        {SA} & 39.2410 & 2.1023 & 1.0829 & 0.5537 & 0.2696 & 0.4243 & 0.5515 \\
        {Average+Contrast} & 39.1985 & 2.0278 & 1.0956 & 0.5397 & 0.2632 & 0.4139 & 0.5375\\
        \textbf{SA+Contrast} & \textbf{40.6113} & \textbf{2.1561} & \textbf{1.0529} & \textbf{0.5601} & \textbf{0.2796} & \textbf{0.4332} & \textbf{0.5572} \\
        \hline
        {AttenAll} & 38.9215 & 2.0271 & 1.0904 & 0.5451 & 0.2578 & 0.4166 & 0.5428 \\
        {SA+CA} & 38.6316 & 2.0414 & 1.0894 & 0.5440 & 0.2579 & 0.4139 & 0.5417 \\
        {SA+NCA} & 39.3278 & 2.0833 & 1.0704 & 0.5490 & 0.2664 & 0.4207 & 0.5459 \\
        \hline
        {SA+Contrast1} & 39.9114 & 2.1006 & 1.0699 & 0.5553 & 0.2731 & 0.4271 & 0.5523 \\
        {SA+Contrast2} & 40.2068 & 2.1115 & 1.0620 & 0.5537 & 0.2725 & 0.4262 & 0.5516 \\
        \hline
        \end{tabularx}
  \end{center}
  \caption{Group captioning performance on the Conceptual Captions and Stock Captions dataset.}
  \label{tab:results}
}
\end{table*}

\section{Comparison with Single Image Captioning}

\begin{figure*}[h!]\vspace{-0.05in}
\begin{center}
\includegraphics[page=1, width=0.99\linewidth]{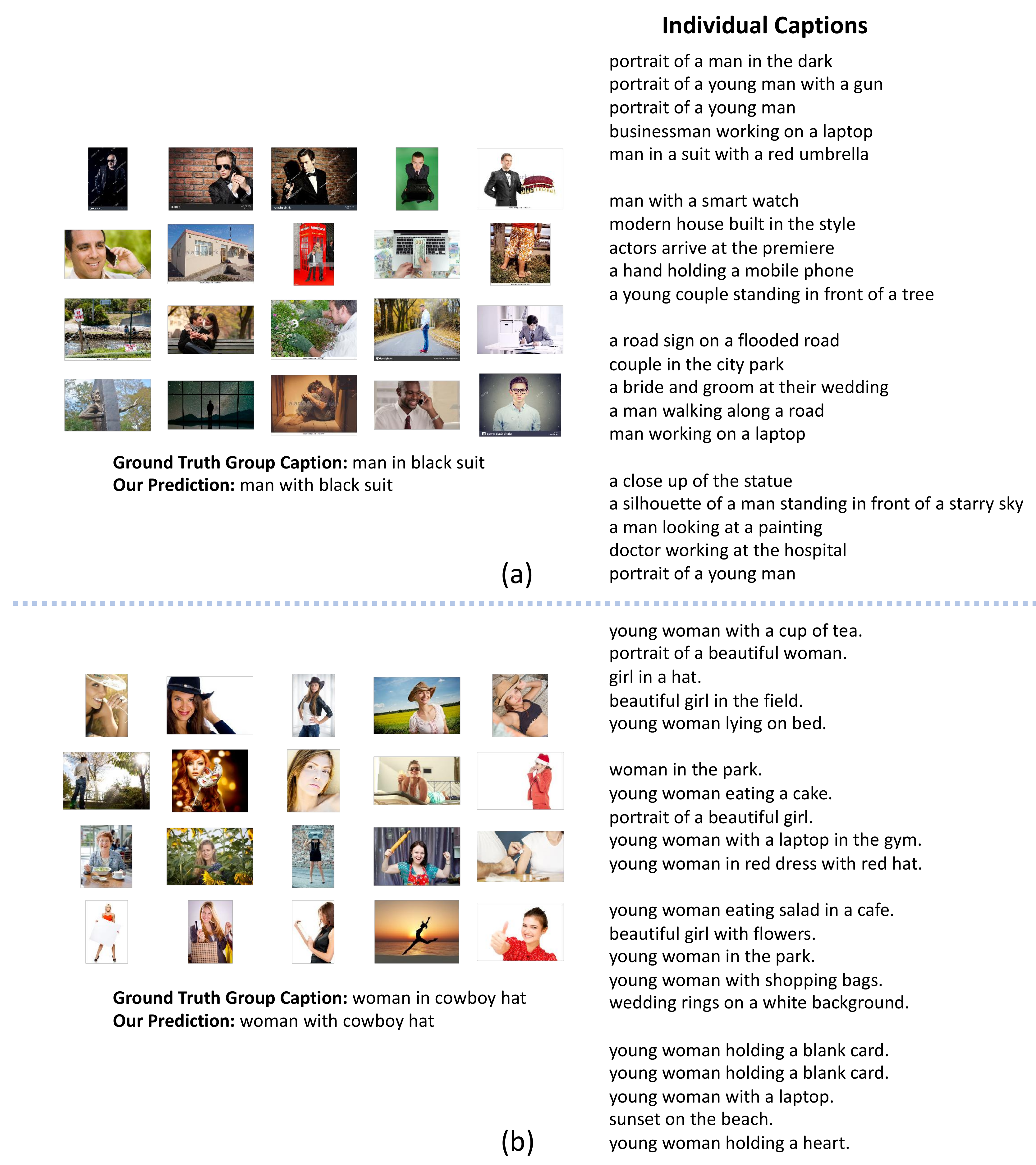}
\end{center}
   \caption{An example on Conceptual Captions dataset to show that the group captioning cannot be easily solved by captioning each image individually. The 20 model-generated captions on the right corresponds to the 20 images on the left in order, where the first 5 are targets and the other 15 are references. In (a), if we are summarizing the 5 target captions on context of reference captions, \texttt{portrait of a man}, which is the most frequent phrase, might be the result, which is not a good description as \texttt{man in black suit}. In (b), if we are summarizing the individual captions to get the group caption, \texttt{young woman} might be the result, which is not as good as \texttt{woman in cowboy hat}. The information needed for group captioning may be missed out in individual captions because the common feature of the group might not be important for individual images. Therefore, captioning images as a group can be more informative. We also perform a limited user study, and most users note that it is almost impossible for them to come up with a summarizing phrase given the individual captions.}\vspace{-0.1in}
\label{fig:20caption}
\end{figure*}

In this section, we describe the difference between our group captioning task and existing individual image captioning task. Captioning each image individually and then summarizing the per-image captions can not solve our task.

Figure \ref{fig:20caption} shows one example from Conceptual Captions and one from Stock Captions. 
The individual image captions are generated using existing image captioning models\footnote{For Conceptual Captions, we use the winning model of Conceptual Captions Challenge Workshop in CVPR2019 to generate captions for each image (\url{https://github.com/ruotianluo/GoogleConceptualCaptioning}). More details of the model can be found at \url{https://ttic.uchicago.edu/~rluo/files/ConceptualWorkshopSlides.pdf}. For Stock Captions, we use the Show, Attend and Tell \cite{xu2015show} captioning model and finetune it on Stock Captions}.
In each figure, the 20 captions on the right corresponds to the 20 images on the left in order, where the first 5 are targets and the other 15 are references.

In (a), while the image group is characterized by \texttt{man in black suit}, the individual captions focus on \emph{man in dark}, \emph{man with a gun}, \emph{portrait of a man}, \emph{man working on a laptop}, etc, thus summarizing them by finding the most frequent phrase will lead to \texttt{portrait of a young man}, which is not a good caption for the image group. In (b), while the image group features for \text{woman in cowboy hat}, individual captions focus on other aspects including \emph{with a cup of tea}(this is an error of the captioning model), \emph{beautiful}, \emph{in the field} or \emph{lying on bed}. Only one per-image caption notices that the woman is \emph{in a hat}. Therefore, if we are summarizing the target per-image captions to get group caption, we will get result \texttt{young woman} or \texttt{beautiful woman}, which miss out the most important feature of the image group (\texttt{woman in cowboy hat}). 
 
 While individual captions might be able to describe each image discriminatively, they does not necessarily include the common properties of the image group, because the common property of the group might not be the significant and distinguishing feature for each image. 
 Therefore, captioning images as a group can capture the information that individual image captioning tend to miss out and thus lead to more informative group captions. Therefore, captioning the group as a whole is different from processing each image individually and then summarizing them. This also explains why merging the image features in early stage using self-attention before generating text descriptions is beneficial.

\section{Analysis of contrastive representation}

\begin{figure*}[h!]\vspace{-0.05in}
\begin{center}
\includegraphics[page=1, width=0.99\linewidth]{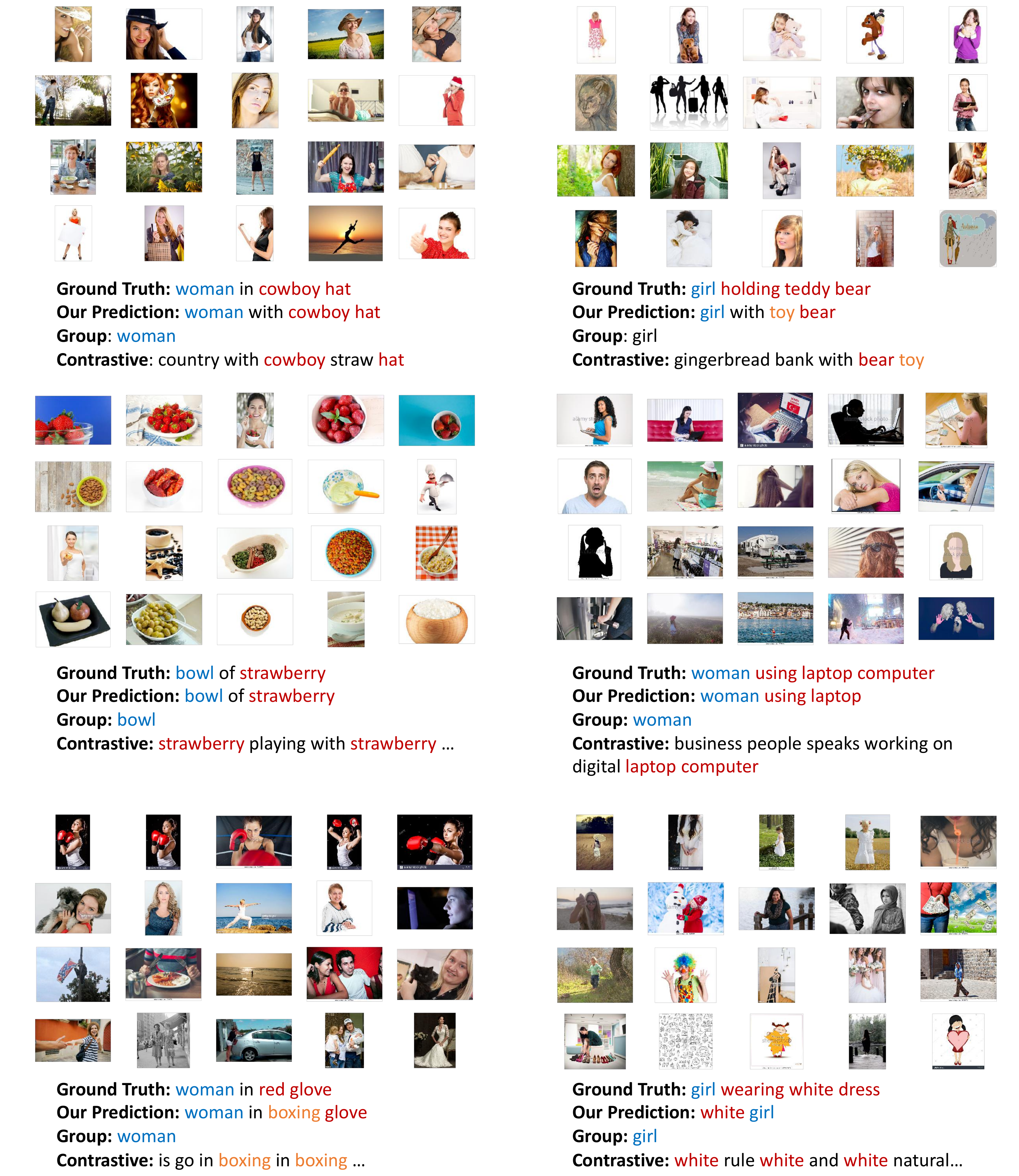}
\end{center}
   \caption{Examples of only using group representation or only using contrastive representation (Corresponding to Table 4 in the main paper). As shown, common information in both image groups (\textcolor{blue}{blue} text) is encoded in the group representation, while the difference between two groups (\textcolor{red}{red} or \textcolor{orange}{orange} text) is in contrastive representation. The first four examples are good cases while the last two examples are failure cases.}\vspace{-0.1in}
\label{fig:example_tab4}
\end{figure*}

In Table 4 of the main paper, we show example results of the captions generated using only group representation or using only contrastive representation. Here in Figure \ref{fig:example_tab4} we show the images of these examples in Table 4. We also provide more examples to illustrate the function of group representation and contrastive representation. The first 3 examples are from Conceptual Captions dataset while the last 3 examples are from Stock Captions. Each example contains of 20 images (four rows), where the first row is target group and the second to fourth rows are reference group.

As shown, the common information in both image groups is encoded in the group representation, while the difference between two image groups is captured by the contrastive representation. The first four examples are good cases while the last two examples are failure cases. In failure case \texttt{woman in red glove}, the contrastive representation fails to capture the red information. In failure case \texttt{girl wearing white dress}, the color white is encoded in the contrastive representation, but its relationship with the girl is wrong in the prediction. 

\section{More Examples}

\begin{figure*}[h!]\vspace{-0.05in}
\begin{center}
\includegraphics[page=1, width=0.99\linewidth]{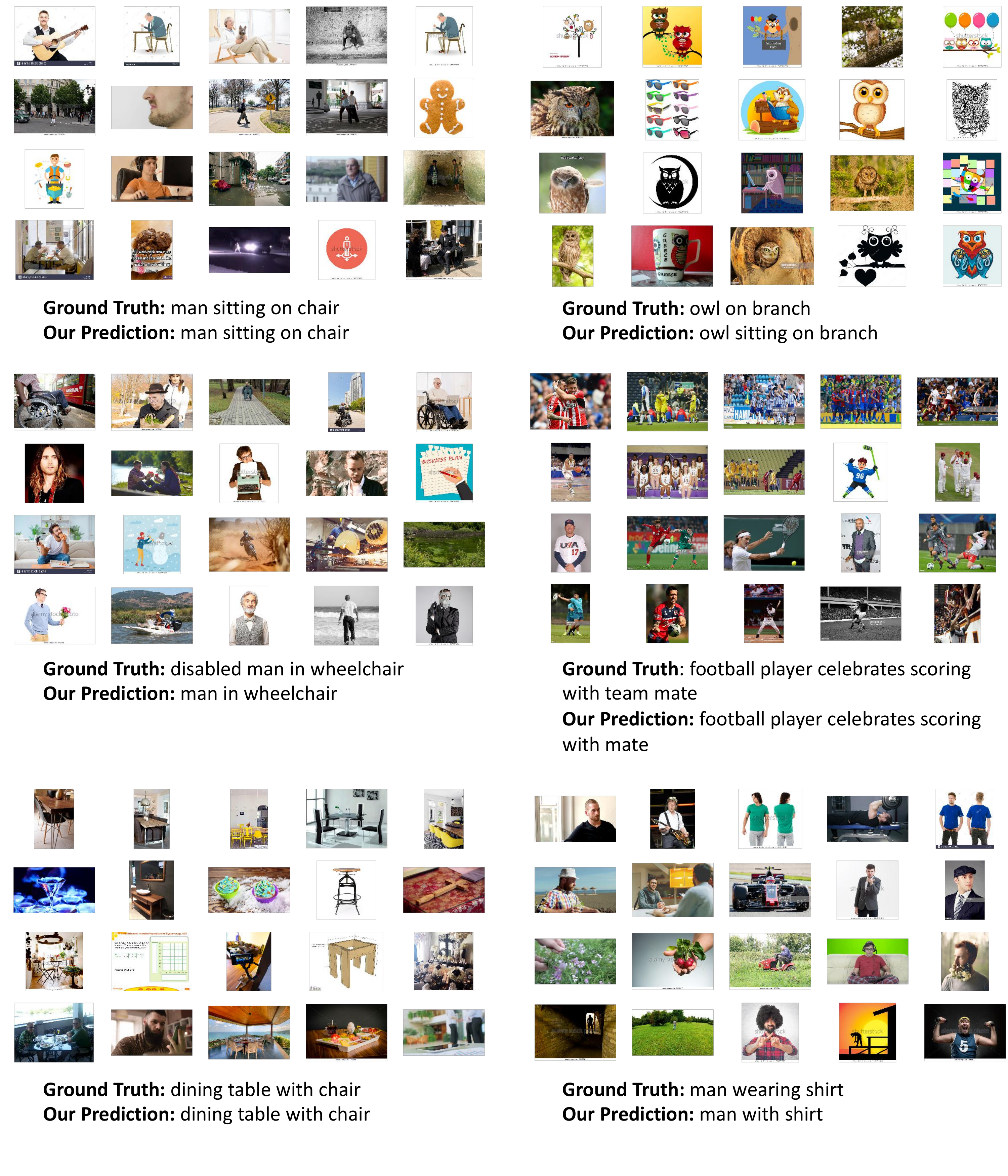}
\end{center}
   \caption{Good examples Conceptual Captions dataset.}\vspace{-0.1in}
\label{fig:example_1}
\end{figure*}

\begin{figure*}[h!]\vspace{-0.05in}
\begin{center}
\includegraphics[page=1, width=0.99\linewidth]{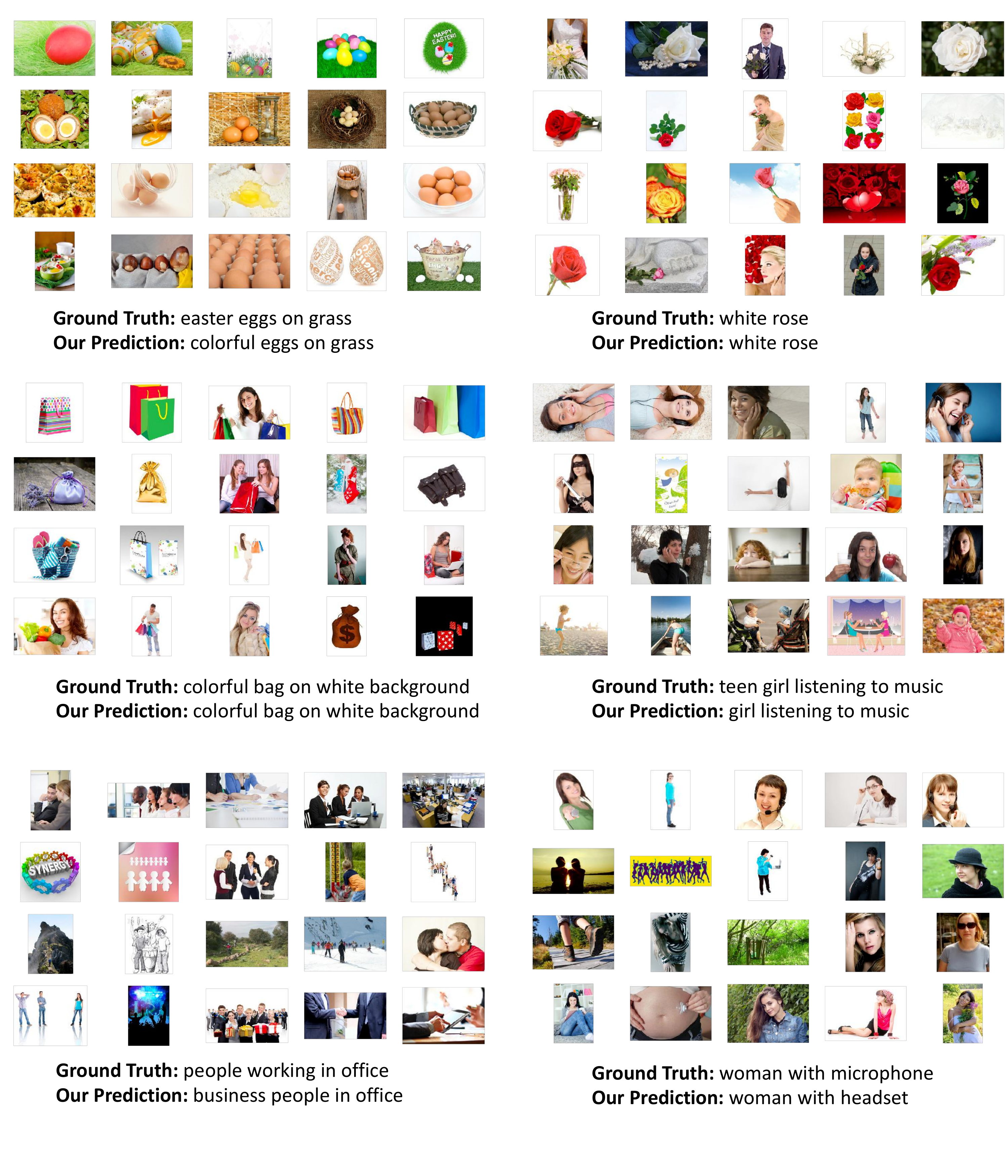}
\end{center}
   \caption{Good examples on Stock Captions dataset.}\vspace{-0.1in}
\label{fig:example_1_stock}
\end{figure*}

\begin{figure*}[h!]\vspace{-0.05in}
\begin{center}
\includegraphics[page=1, width=0.99\linewidth]{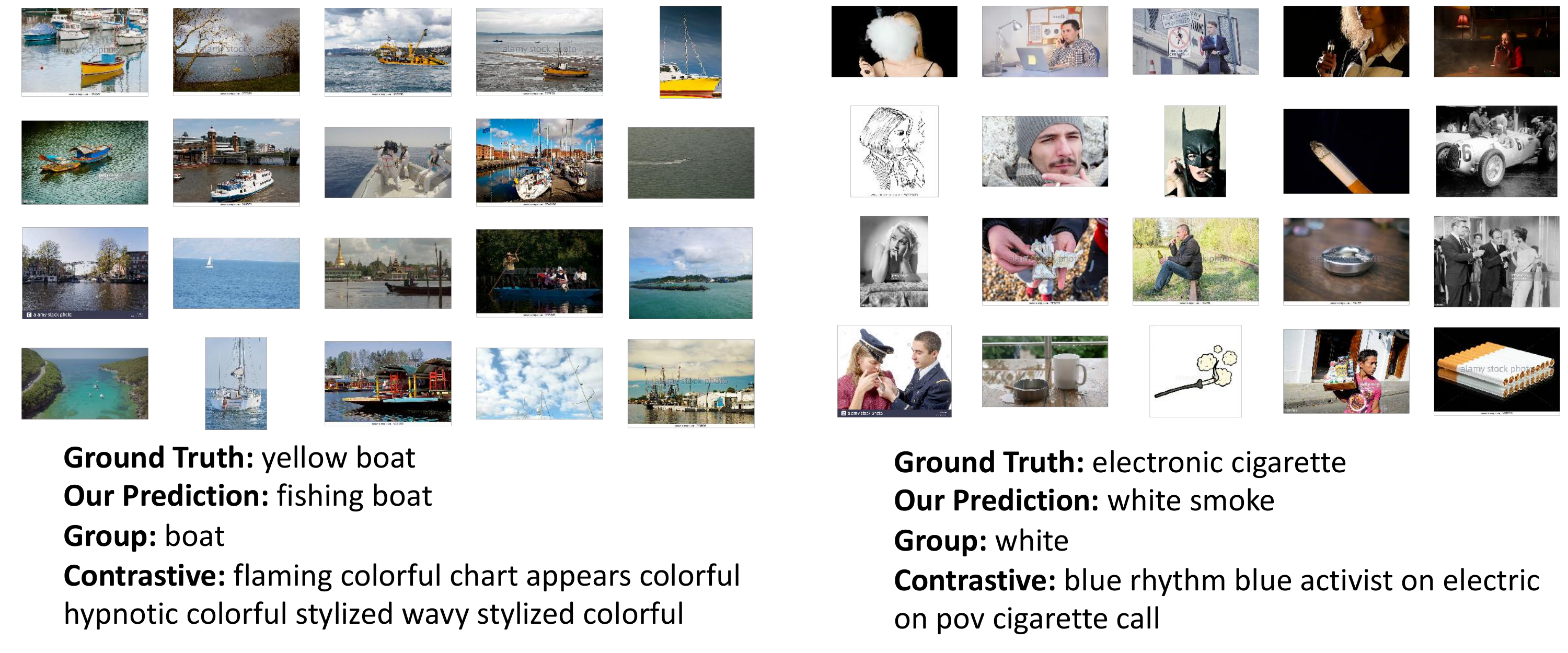}
\end{center}
   \caption{Failure cases on Conceptual Captions dataset. For the first example, the model predicts \texttt{fishing boat} instead of \texttt{yellow boat}, which is less discriminative. This may be because the model does not capture features of the small boat well. For the example on the right, the model prediction (\texttt{white smoke}) may be dominated by one dominant image in the target group.}\vspace{-0.1in}
\label{fig:example_3}
\end{figure*}

\begin{figure*}[h!t]\vspace{-0.05in}
\begin{center}
\includegraphics[page=2, width=0.99\linewidth]{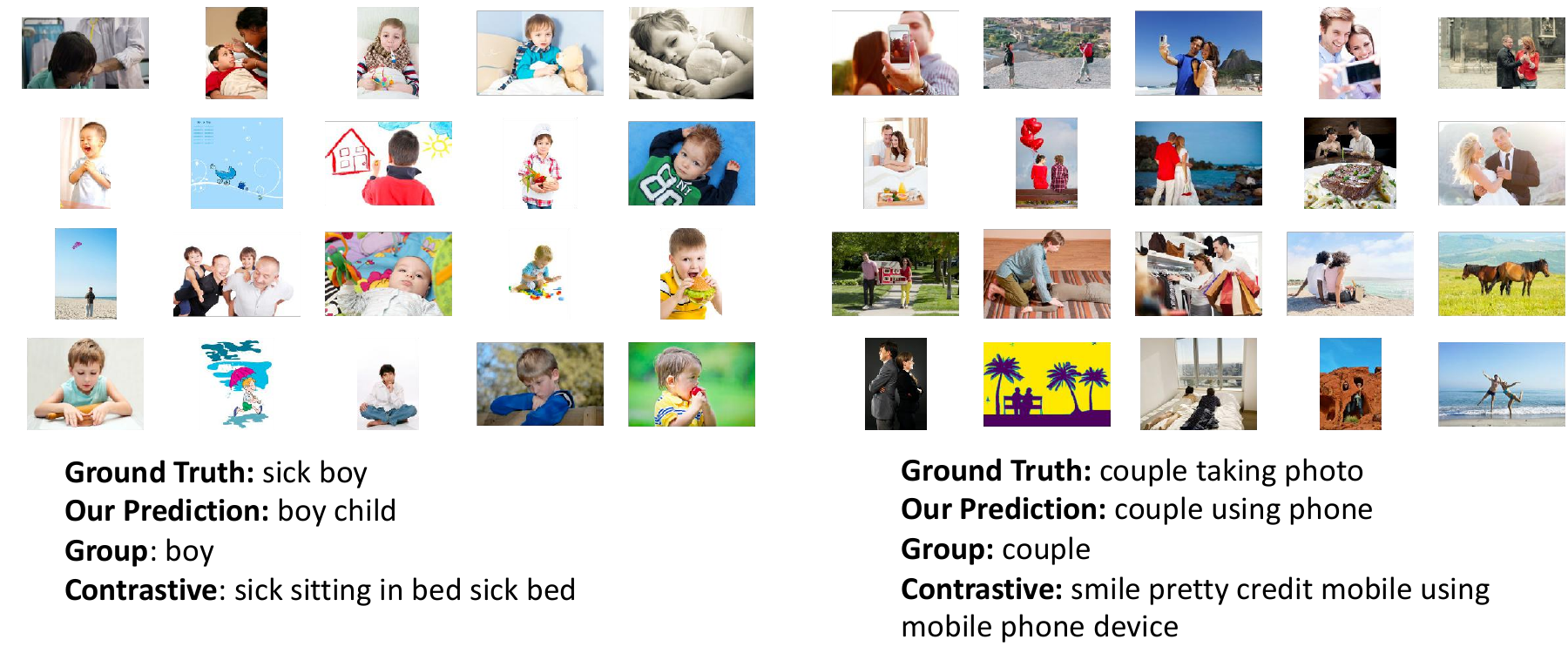}
\end{center}
   \caption{Failure cases on Stock Captions dataset. For the first example, the model prediction does not notice that the boy is \texttt{sick}. We further look into the model output when using only the group representation or contrastive representation, where the \emph{sick} information is captured in the contrastive representation, but may not be strong enough to be decoded out in the prediction. For the second example, the model prediction is correct but not as good as groundtruth. }\vspace{-0.1in}
\label{fig:example_3_stock}
\end{figure*}

Figure \ref{fig:example_1}  and Figure \ref{fig:example_1_stock} show more good examples on Conceptual Captions and Stock Captions respectively. Figure \ref{fig:example_3} and Figure \ref{fig:example_3_stock} show failure cases on the two datasets respectively. Similar as above, in each example, the first row is target group while the other rows are reference group. Analysis for the failure cases (Figure \ref{fig:example_3}, Figure \ref{fig:example_3_stock}) can be found in the captions of each figure.

\end{document}